\documentclass[10pt,twocolumn,letterpaper]{article}

 \usepackage[pagenumbers]{cvpr} 

\definecolor{cvprblue}{rgb}{0.21,0.49,0.74}
\usepackage[pagebackref,breaklinks,colorlinks,allcolors=cvprblue]{hyperref}

\def\confName{CVPR}
\def\confYear{2027}

\definecolor{bestgreen}{HTML}{DCECDF}
\definecolor{improvegray}{HTML}{808080}
\definecolor{noimprovegray}{HTML}{C4C4C4}
\definecolor{deltagain}{HTML}{557F3E}
\definecolor{deltaloss}{HTML}{D62728}
\definecolor{deltazero}{HTML}{B0B0B0}

\providecommand{\ReBindDeltaSize}{5.0}
\providecommand{\ReBindDeltaLeading}{5.8}

\providecommand{\ReBindBest}[1]{}
\renewcommand{\ReBindBest}[1]{%
  \begingroup
  \setlength{\fboxsep}{0.6pt}%
  \colorbox{bestgreen}{\strut #1}%
  \endgroup
}

\providecommand{\ReBindDelta}[3]{}
\renewcommand{\ReBindDelta}[3]{%
  \begingroup
  \fontsize{\ReBindDeltaSize}{\ReBindDeltaLeading}\selectfont
  \textcolor{#1}{\ensuremath{#2}#3}%
  \endgroup
}

\usepackage{amsmath}
\usepackage{amssymb}
\usepackage{graphicx}
\usepackage{array}
\usepackage{tabularx}
\usepackage{algorithm}
\usepackage{algpseudocode}
\algrenewcommand\algorithmicrequire{\textbf{Input:}}
\algrenewcommand\algorithmicensure{\textbf{Output:}}

\usepackage{booktabs}   
\usepackage{multirow}   
\usepackage{makecell}   
\usepackage{adjustbox}  
\usepackage[table]{xcolor} 
\usepackage{colortbl}  
\title{When Instruction-Visual Binding Weakens: Diagnosing and Mitigating Hallucinations in Large Vision-Language Models}

\author{
  Jiahui Hu\textsuperscript{1},
  Bo Liu\textsuperscript{2},
  Xiaolong Luo\textsuperscript{3},
  Yuan Xie\textsuperscript{2},
  Dan Liu\textsuperscript{1},
  Huamin Wang\textsuperscript{1},
  Fujun Han\textsuperscript{2}\thanks{Corresponding author.}\\[2pt]
  \textsuperscript{1}Southwest University\quad
  \textsuperscript{2}The Chinese University of Hong Kong, Shenzhen\\
  \textsuperscript{3}City University of Hong Kong\\
  {\small\ttfamily hujiahui@email.swu.edu.cn, hanfujun@cuhk.edu.cn}
}

\begin{document}
\maketitle
\begin{abstract}
Large vision-language models (LVLMs) have demonstrated remarkable performance on multimodal tasks, yet hallucinations continue to severely limit their reliability. Existing gradient-based methods evaluate the contributions of different token categories separately, overlooking their intrinsic relationships and how they evolve during generation. To bridge this gap, we introduce a novel hallucination diagnostic metric called the Instruction--Visual Binding Score (IVBS), which quantifies the visual sensitivity of the current query's dependence on the instruction. Based on the proposed IVBS, we reveal an insightful finding, \textit{i.e.}, IVBS declines typically coincide with hallucination generation, indicating weakened instruction--visual binding. Motivated by this observation, we propose ReBind, an effective and training-free inference framework with two sequential stages, to mitigate hallucinations. Specifically, we first design Instruction--Visual Semantic Enhancement (IVSE) to strengthen instruction--visual representations for reuse in subsequent generation during the prefill stage. Then, in the decoding stage, we introduce Binding-Guided Rethinking (BGR) to correct token choices at risk of hallucination. Furthermore, to provide visual evidence for subsequent predictions, we design a Prefill Memory Reinjection (PMR) method. Finally, extensive experiments and comparisons are conducted on multiple dominant benchmarks, \textit{e.g.}, POPE, CHAIR, MME, and AMBER. The comprehensive experimental results demonstrate that, compared with current works, our ReBind shows promising performance and can significantly mitigate hallucinations in LVLMs, which verifies the effectiveness of the proposed frameworks.
\end{abstract}    
\section{Introduction}
\label{sec:intro}

Large Vision-Language Models (LVLMs) have achieved great progress in tasks, \textit{e.g.}, image captioning and visual question answering, by integrating visual encoders with large language models \cite{wang2024qwen2,chen2023shikra,liu2023visual,liu2024improved,zhu2024minigpt,dai2023instructblip,SOU,OVID}. However, LVLMs remain prone to hallucinations \cite{zechen2024hallucination,farquhar2024detecting,liu2024survey,min2025mitigating}, \textit{i.e.}, generating content inconsistent with or absent from the input image, which not only undermines the output trustworthiness but also restricts their deployment in safety-critical scenarios \textit{e.g.}, autonomous driving~\cite{wang2026rac3,chen2024driving, MLLMISU}, medical imaging~\cite{hu2024omnimedvqa,hu2023advancing}, and embodied intelligence~\cite{xiao2026dive}. Thus, mitigating hallucinations has become a critical issue for enhancing the trustworthiness and practicality of LVLMs.

\begin{figure}[t]
    \centering
    \includegraphics[width=\columnwidth]{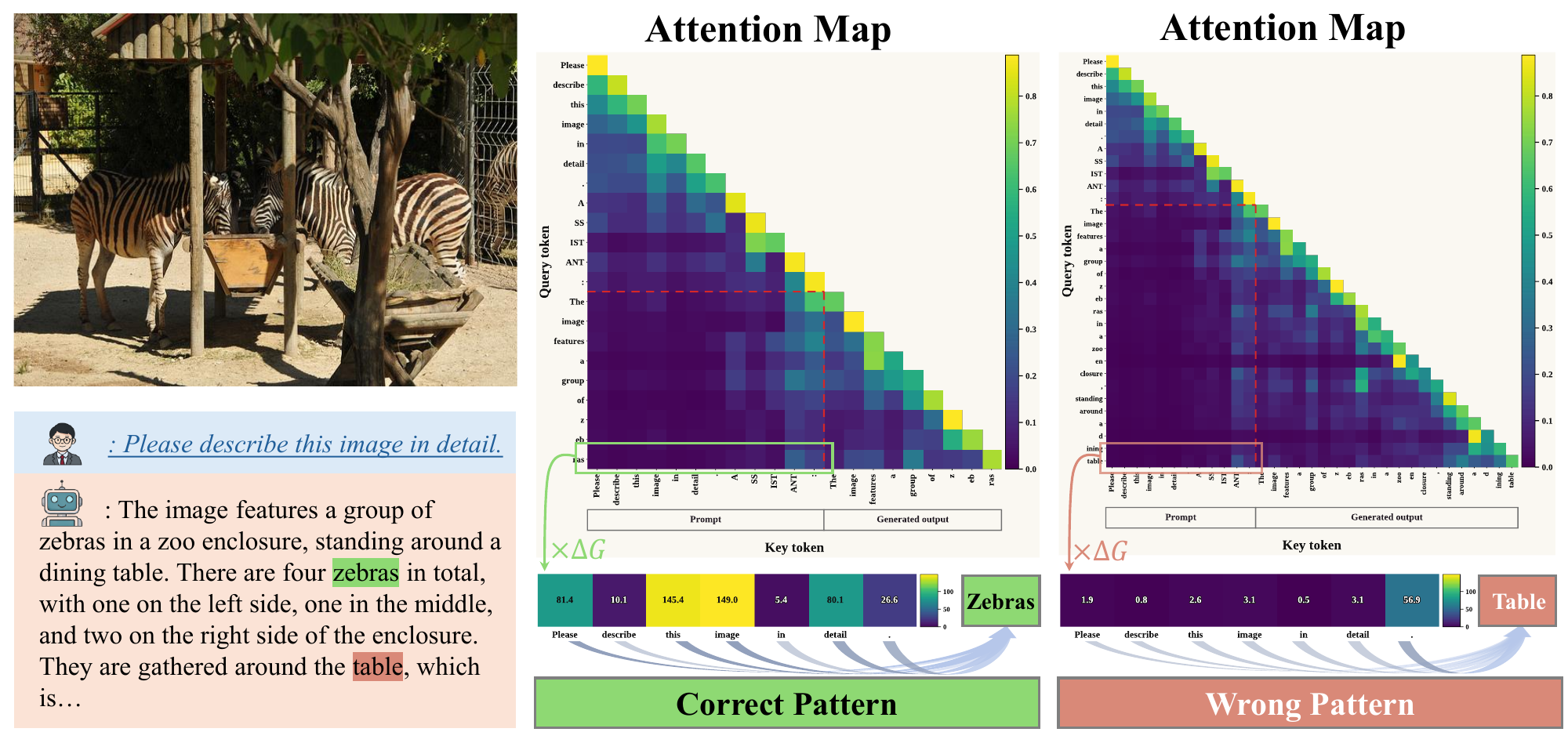}
    \caption{The illustration of instruction--visual binding. We can find that attention and saliency maps for faithful and hallucinated tokens. Attention patterns remain similar, but instruction-region saliency weakens for the hallucinated token, indicating reduced instruction--visual binding.}
    \label{fig:ivbs_token}
    \vspace{-8pt} 
\end{figure}

\begin{figure*}[t]
    \centering
    \begin{minipage}[t]{0.48\textwidth}
        \centering
        \includegraphics[width=\linewidth]{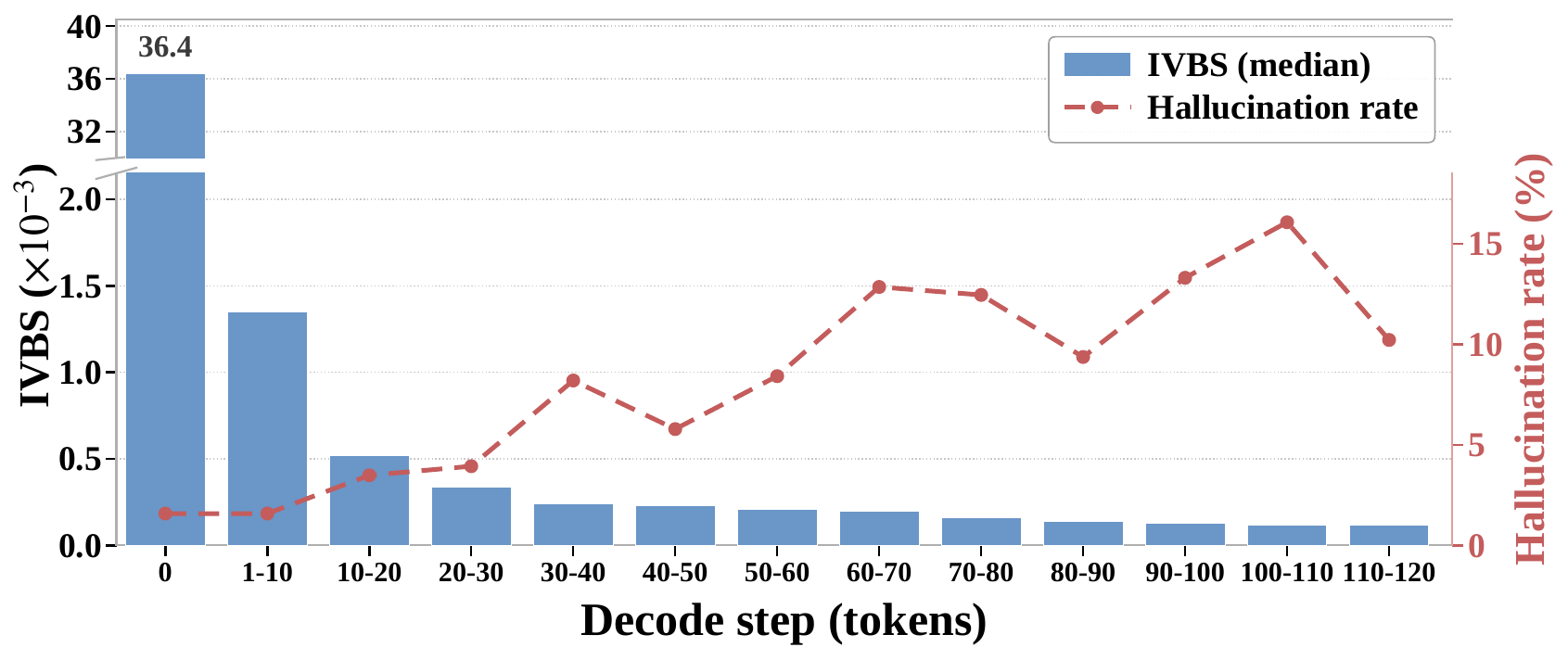}
    \caption{Binding dynamics during caption generation with LLaVA-1.5-7B on COCO 2014. Bars show median IVBS; the dashed curve shows hallucination rate. Binding peaks at the first step and rapidly declines, while hallucination generally increases.}
        \label{fig:ivbs_steps}
    \end{minipage}
    \hfill
    \begin{minipage}[t]{0.48\textwidth}
        \centering
        \includegraphics[width=\linewidth]{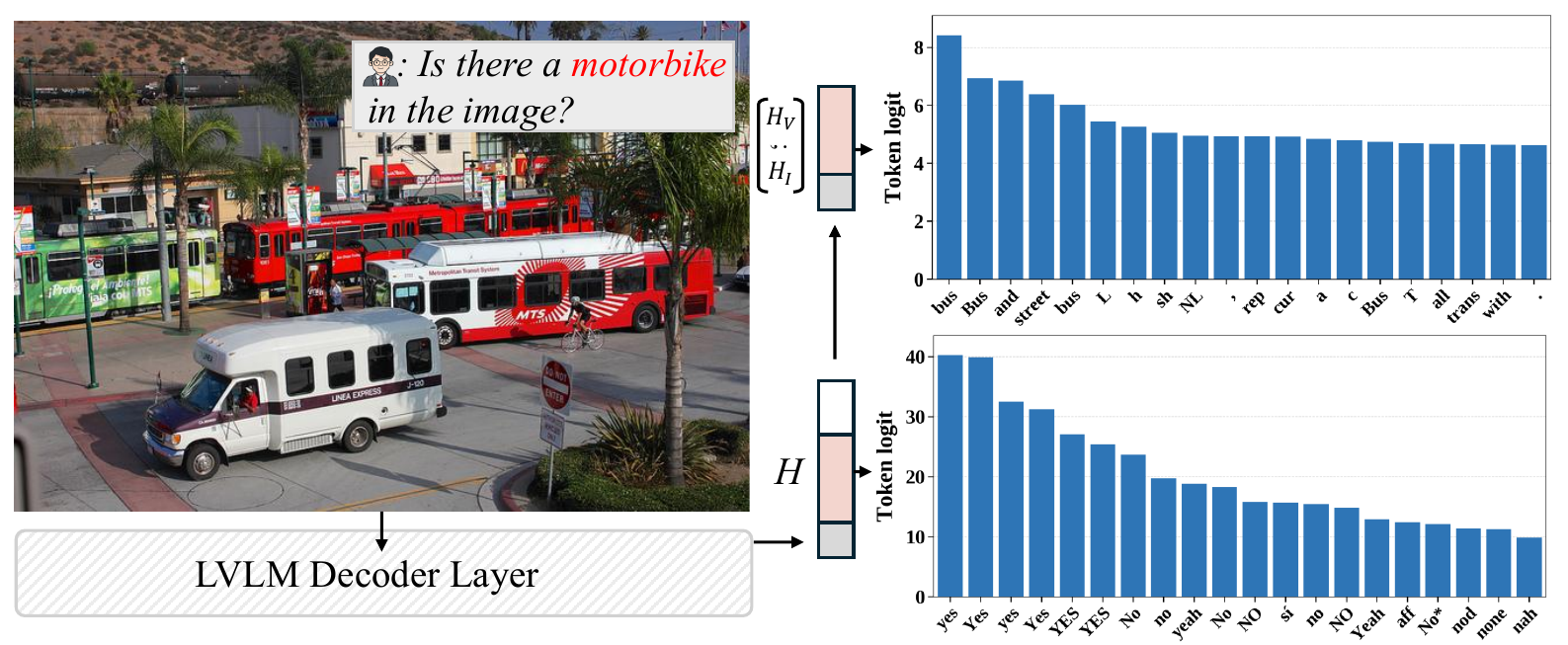}
        \caption{Logit-lens analysis of first-token prediction. Top: the visual--instruction subset reveals image-consistent semantics including ``bus.'' Bottom: the prediction position favors the incorrect answer despite internally available visual information.}
        \label{fig:logit_lens}
    \end{minipage}
    \vspace{-8pt}
\end{figure*}

To mitigate hallucinations in LVLMs, existing research can mainly be divided into three categories, \textit{i.e.}, retraining-based~\cite{sun2024aligning,gunjal2024detecting,liu2024mitigating,yu2024rlhf}, auxiliary-model-based~\cite{yin2024woodpecker,deng2024seeing}, and training-free methods~\cite{leng2024mitigating,huang2024opera,zou2024look}. 
Owing to their ability to operate without modifying model parameters, training-free approaches have attracted increasing attention for mitigating hallucinations in LVLMs~\cite{wan2025only,liu2025reducing,li2025hidden}. These methods typically adopt contrastive decoding~\cite{leng2024mitigating,wang2024mitigating,chuang2024dola,favero2024multi} or directly manipulate attention~\cite{huang2024opera,sarkar2025mitigating,tang2025seeing,liu2024paying} to weaken language priors, yet they still lack interpretability regarding the causes of hallucinations in autoregressive models. Attention maps primarily capture attention weight distributions rather than the sensitivity of predictions to input tokens, making it difficult to distinguish correct from hallucinated outputs, as illustrated in Fig.~\ref{fig:ivbs_token}. Face the challenge, recent works on gradient-aware signals offer a new perspective by characterizing how input tokens contribute to final predictions~\cite{zhang2026hallucination,wang2026mitigating}. Despite these works broadening the diagnostic lens, existing gradient-based explorations still face two key limitations, \textit{i.e.}, (1)~Neglect of cross-modal coupling. Existing methods primarily compute the gradient contributions of visual or instruction tokens in isolation, overlooking their intrinsic connection, namely \textit{visually modulated instruction dependency}. (2)~Lack of a long-horizon dynamic perspective. The dynamic of gradient-based metrics during long-form generation remain underexplored. 

To address the limitations, we first combine attention with visual-differential gradients to propose a novel hallucination diagnostic metric, the Instruction--Visual Binding Score (IVBS), which quantifies the visual sensitivity of the current query's dependency on instructions. Our analysis first reveals a significant correlation between declining IVBS and hallucination occurrence (as shown in Fig.~\ref{fig:ivbs_token}). Furthermore, we find that hidden representations at the prefill stage contain the strongest instruction-relevant visual information (Fig.~\ref{fig:ivbs_steps}); even so, hallucinations still occur at the initial generation step. Subsequent in-depth observation uncovers an intriguing phenomenon, \textit{i.e.}, question-relevant visual evidence is already encoded within the instruction--visual subset of the prefill hidden representations, yet the final answer remains incorrect (Fig.~\ref{fig:logit_lens}).\par

Building on these observations, we then propose ReBind (Fig.~\ref{fig:framework}), a training-free, two-stage inference intervention method designed to mitigate hallucinations caused by weakened instruction--visual binding. Specifically, during decoding, when the current candidate token's IVBS is monitored to fall below a dynamic threshold, Binding-Guided Rethinking (BGR) performs local replacement correction; considering that instruction-relevant visual information may have decayed during long-form generation, we introduce Prefill Memory Reinjection (PMR) upon token replacement to restore input evidence for the next prediction; meanwhile, as initial generation may still suffer from hallucinations due to representation dispersion, we propose Instruction--Visual Semantic Enhancement (IVSE) at the prefill stage to refine instruction--visual memory and guide the starting point of the answer. Through extensive experiments, our method demonstrates substantial hallucination mitigation performance across different LVLMs on multiple benchmarks, validating its effectiveness. \par

In summary, our contributions are as follows:
\begin{itemize}
    \item \textbf{Insightful Binding Analysis}. We revisit hallucinations from the perspective of visual-differential gradients and propose a novel diagnostic metric, IVBS, to quantify visually modulated instruction dependency. Through systematic analysis, we reveal the intrinsic relationship between declining IVBS and hallucination generation.
    
    \item \textbf{Effective Two-Stage Intervention Framework}. We propose ReBind, an effective, training-free, inference intervention framework with three well-designed methods, \textit{i.e.}, Instruction–Visual Semantic Enhancement (IVSE), Binding-Guided Rethinking (BGR), and Prefill Memory Reinjection (PMR), to strengthen and reuse prefill visual-instruction representations. 
    \item \textbf{Extensive Experimental Validation}. We conduct extensive experiments across multiple benchmarks and LVLM architectures. The experimental results show that, compared with conventional decoding, ReBind improves POPE F1 by an average of 10.17\% and reduces the CHAIR instance-level hallucination rate by an average of 35.33\% on LLaVA-1.5 and InstructBLIP.
\end{itemize}

\section{Related Work}

\noindent\textbf{Contrastive Decoding and Attention Manipulation}. Existing training-free methods for mitigating hallucinations in LVLMs mainly fall into contrastive decoding and attention manipulation. The former suppresses unreliable outputs by contrasting prediction distributions under different conditions, \textit{e.g.}, VCD~\cite{leng2024mitigating} compares distributions conditioned on original and distorted images, ICD~\cite{wang2024mitigating} introduces instruction perturbations, and DoLa~\cite{chuang2024dola} contrasts logits from deeper and shallower layers to emphasize factual knowledge. The latter improves visual information utilization through attention and representation interventions, \textit{e.g.}, OPERA~\cite{huang2024opera} uses penalties and backtracking to reduce reliance on summary tokens. ONLY~\cite{wan2025only} performs single-layer intervention based on a text-to-visual entropy ratio. AGLA~\cite{an2025mitigating} integrates global and local cues. MemVR~\cite{zou2024look} reinjects visual memory to supplement evidence. VISTA~\cite{li2025hidden} combines representation steering with intermediate-layer logit augmentation. Despite their progress, these methods offer limited explanations of hallucination causes during autoregressive generation.

\noindent\textbf{Gradient-Based Hallucination Mitigation}. Recently, some gradient-based studies provide a new perspective. LLaVA-Saliency~\cite{zhang2026hallucination} combines attention and gradients to assess the influence of historical tokens, while GACD~\cite{wang2026mitigating} evaluates token contributions using first-order Taylor gradients to rebalance cross-modal information. However, existing investigations still overlook instruction--visual coupling and lack a long-horizon dynamic perspective. In contrast, our method addresses these challenges by using IVBS to reveal the association between weakened instruction--visual binding and hallucination generation, motivating ReBind, a two-stage inference framework. ReBind employs IVSE to strengthen instruction--visual representations during prefill, BGR to correct token choices when binding responses are low, and PMR to reintroduce reference memory for subsequent predictions. Together, these components mitigate binding degradation and achieve strong performance.
\section{Observation and Analysis}
\label{sec:observations}

\begin{figure*}[t] 
    \centering
    \includegraphics[width=0.9\textwidth]{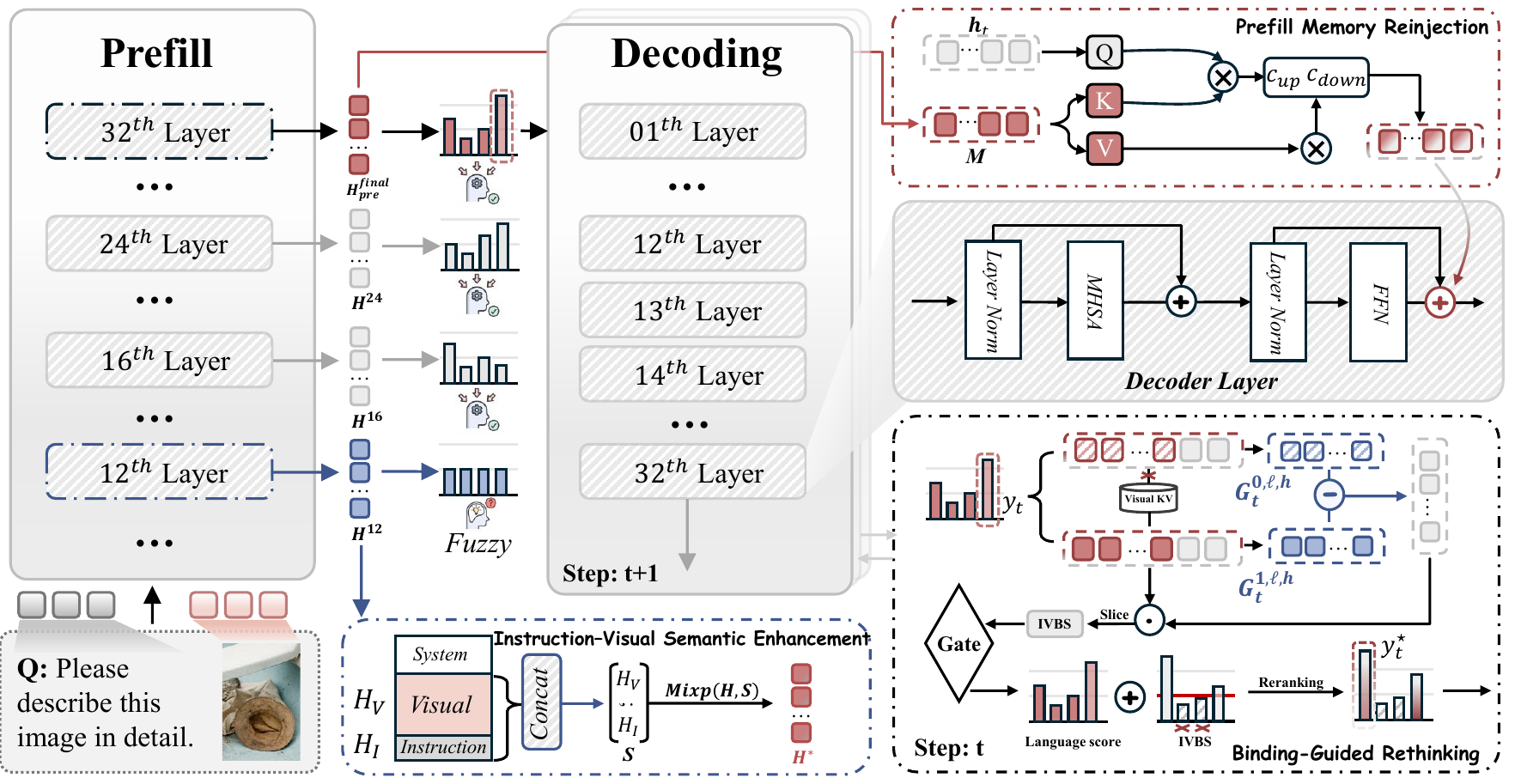}
    \caption{Overview of ReBind. During prefill, IVSE strengthens instruction--visual representations according to prediction uncertainty, and the final prefill representations are cached as reference memory. During decoding, BGR uses IVBS to correct token choices with weak binding responses, and PMR reinjects the cached memory after token replacement to support subsequent predictions.}
    \label{fig:framework}
    \vspace{-8pt} 
\end{figure*}

\subsection{Token-Level Analysis of the IVBS}
\label{sec:ivbs}
To quantify the visual sensitivity of the current query's dependence on instructions, we propose an attention analysis framework based on visual-differential gradients. Given a multimodal context $\mathbf{x}_t$ of length $n_t$ comprising visual input, user instructions, and the generated prefix, model $\mathcal{M}$ outputs logits $\mathbf{z}_t$, probabilities $\mathbf{p}_t$, and attention matrices $\{\mathbf{A}_t^{\ell,h}\}_{\ell=1,h=1}^{L,H}$, where $L$ and $H$ denote the numbers of layers and heads. We analyze the current query's last attention row and index instruction positions by $\mathcal{I}$.

We construct two local scoring conditions: \textit{visual retention} ($\alpha=1$) preserves visual keys and values, whereas \textit{visual suppression} ($\alpha=0$) zeroes them in selected scoring layers without changing the context or sequence positions. Both conditions target the standard multimodal prediction $\hat{y}_t=\arg\max_{v\in\mathcal{W}}p_t(v)$, with loss $\mathcal{L}_t^\alpha=-\log p_t^\alpha(\hat{y}_t)$, where $p_t^\alpha$ is the distribution under condition $\alpha$.

We introduce differentiable multiplicative coefficients $\boldsymbol{\theta}_t^{\ell,h}$, initialized to $\mathbf{1}$, on the last attention row. For efficiency, local backpropagation is restricted to the selected scoring layers:
\begin{equation}
    \mathbf{G}_t^{\alpha,\ell,h}
    = \frac{\partial\mathcal{L}_t^\alpha}{\partial\boldsymbol{\theta}_t^{\ell,h}},
\end{equation}
\begin{equation}
    \Delta\mathbf{G}_t^{\ell,h}
    = \mathbf{G}_t^{1,\ell,h}-\mathbf{G}_t^{0,\ell,h}.
\end{equation}
The difference captures changes in each connection's gradient response to the same target under the two visual conditions. We obtain visual-differential saliency by multiplying the visual-retention attention row $\mathbf{a}_t^{1,\ell,h}$ by the differential gradient element-wise and averaging absolute values across heads:
\begin{equation}
    \mathbf{s}_t^\ell
    = \frac{1}{H}\sum_{h=1}^{H}
    \left|\mathbf{a}_t^{1,\ell,h}\odot\Delta\mathbf{G}_t^{\ell,h}\right|,
\end{equation}
where $\odot$ denotes element-wise multiplication. We then $\ell_2$-normalize the entire saliency row, average over instruction positions, and take the minimum across scoring layers $\mathcal{L}_{\mathrm{score}}$:
\begin{equation}
    \bar{\mathbf{s}}_t^\ell
    = \frac{\mathbf{s}_t^\ell}{\max\left(\|\mathbf{s}_t^\ell\|_2,\epsilon\right)},
\end{equation}
\begin{equation}
    \operatorname{IVBS}_t
    = \min_{\ell\in\mathcal{L}_{\mathrm{score}}}
    \frac{1}{|\mathcal{I}|}\sum_{j\in\mathcal{I}}\bar{s}_{t,j}^\ell.
\end{equation}
Here, $\epsilon$ ensures numerical stability. Lower IVBS indicates a weaker visual-differential response over instruction positions for the current prediction, providing an interpretable binding signal. As shown in Fig.~\ref{fig:ivbs_token}, hallucinated tokens exhibit weaker saliency and lower IVBS than faithful tokens, linking weakened binding to hallucinations.

\subsection{Analysis of IVBS Across Generation Steps}
\label{sec:binding_steps}
We examine IVBS across generation steps using LLaVA-1.5-7B~\cite{liu2024improved} captions for 500 randomly sampled COCO 2014 images~\cite{lin2014microsoft}, prompted with \textit{``Please describe this image in detail.''} We compute cumulative IVBS and hallucination rates within generation-step intervals. As shown in Fig.~\ref{fig:ivbs_steps}, IVBS rapidly declines while hallucination rate increases, supporting the binding--hallucination association in Section~\ref{sec:ivbs}. Within the analyzed range, first-token IVBS markedly exceeds subsequent values, even their cumulative sum. The absence of output history and shorter sequence distances to image and instruction tokens may help preserve stronger initial binding. This motivates reusing prefill representations as an instruction--visual anchor to supplement visual evidence during subsequent generation and reduce overreliance on language priors.

\subsection{Analysis of First-Token Prediction}
\label{sec:logit_lens}
Although IVBS is substantially higher at the first generation step than at subsequent steps, hallucinations can still occur at this stage. To further investigate this phenomenon, we use logit lens~\cite{li2025hidden} to project and analyze hidden representations across decoder layers during first-token prediction. We observe that even when the representation at the prediction position favors an incorrect answer, image-consistent and task-relevant visual information remains decodable from the visual--instruction subset at the same layer. As shown in Fig.~\ref{fig:logit_lens}, when determining whether an image contains a motorcycle, the prediction representation favors the incorrect answer \textit{\textquotedblleft{}Yes\textquotedblright{}}, whereas the visual--instruction subset still reveals semantics such as \textit{\textquotedblleft{}bus\textquotedblright{}} that match the actual image content. This indicates that relevant visual information remains within the model when hallucinations occur, but may not be fully integrated and used for answer prediction. This finding motivates us to revisit the model's existing instruction--visual representations and explore how to strengthen and reuse their relevant semantics so that they contribute more fully to answer prediction.
\section{Methodology}
\label{sec:method}
Motivated by the preceding observations, we propose ReBind (Fig.~\ref{fig:framework}), a training-free inference-time intervention framework comprising two sequential stages: prefill and decoding. During prefill, Instruction--Visual Semantic Enhancement (IVSE) strengthens instruction--visual semantics based on prediction uncertainty. During decoding, Binding-Guided Rethinking (BGR) uses IVBS to locally correct token choices when binding responses are low. Prefill Memory Reinjection (PMR) caches the final prefill representations as reference memory and reinjects them after BGR-triggered token replacement to provide instruction--visual information for subsequent predictions.

\subsection{Instruction--Visual Semantic Enhancement}
\label{sec:ivse}

Despite the stronger binding response at first-token prediction, hallucinations can still occur at this stage. Inspired by our layer-wise representation analysis and MemVR~\cite{zou2024look}, we select enhancement layers according to prediction uncertainty during prefill and integrate relevant semantics from the visual--instruction subset into the current representations.

Specifically, we map the last-position representation at layer $\ell$ to logits $\mathbf{z}^{(\ell)}$ through the output normalization and output head, and compute the normalized top-10 entropy:
\begin{equation}
    \mathbf{q}^{(\ell)}
    =
    \operatorname{softmax}
    \left(
    \operatorname{TopK}(\mathbf{z}^{(\ell)},10)
    \right),
\end{equation}
\begin{equation}
    U^{(\ell)}
    =
    -\frac{1}{\log 10}
    \sum_{i=1}^{10}
    q_i^{(\ell)}\log q_i^{(\ell)}.
\end{equation}
When $U^{(\ell)}$ exceeds a threshold $\tau$, we enhance the corresponding layer representations $\mathbf{H}^{(\ell)}\in\mathbb{R}^{N\times d}$. We extract representations at visual positions $\mathcal{V}$ and user instruction positions $\mathcal{I}$, and concatenate them along the token dimension to form the enhancement source:
\begin{equation}
    \mathbf{S}^{(\ell)}
    =
    [\mathbf{H}_{\mathcal{V}}^{(\ell)};
     \mathbf{H}_{\mathcal{I}}^{(\ell)}].
\end{equation}

To integrate these subset semantics into the current layer, we define a memory adapter using $\mathbf{S}^{(\ell)}$ as its memory source. For hidden representations $\mathbf{X}$ and a memory source $\mathbf{B}$, the adapter uses $\mathbf{B}$ itself as both projection factors:
\begin{equation}
    \mathcal{A}(\mathbf{X},\mathbf{B})
    =
    c_{\mathrm{up}}c_{\mathrm{down}}
    \left(\mathbf{X}\mathbf{B}^\top\right)\mathbf{B}.
\end{equation}
Here, $c_{\mathrm{up}}$ and $c_{\mathrm{down}}$ are the ratios of the mean absolute magnitudes of the final-layer MLP's up- and down-projection weights to that of $\mathbf{B}$, respectively, providing magnitude calibration. We then match the adapter output's magnitude to the original representations and interpolate between them:
\begin{equation}
    \mathrm{Mix}_{r}(\mathbf{X},\mathbf{B}) = (1-r)\mathbf{X} + r \frac{\langle|\mathbf{X}|\rangle}{\max(\langle|\mathcal{A}(\mathbf{X},\mathbf{B})|\rangle,\epsilon)} \mathcal{A}(\mathbf{X},\mathbf{B}).
\end{equation}
Here, $\langle|\cdot|\rangle$ denotes the mean absolute element value, $r$ controls the mixing strength, and $\epsilon$ ensures numerical stability.

Using the full sequence $\mathbf{H}^{(\ell)}$ as both the query and residual, we incorporate the subset semantics into the current layer representations:
\begin{equation}
    \mathbf{H}^{(\ell),\star}
    =
    \operatorname{Mix}_{\beta}
    \left(
    \mathbf{H}^{(\ell)},\mathbf{S}^{(\ell)}
    \right).
\end{equation}
Here, $\beta$ controls the enhancement strength. The enhanced representations propagate through subsequent decoder layers, integrating their influence into the final prefill output. PMR caches and reuses this final output rather than the intermediate enhancement.

\subsection{Binding-Guided Rethinking}

Based on the preceding observations, BGR evaluates instruction--visual binding before committing a token. Rethinking is triggered when the greedy token's IVBS falls below a fraction of the historical mean:
\begin{equation}
    \mu_t
    =
    \frac{1}{|\mathcal{H}_t|}
    \sum_{b\in\mathcal{H}_t}b,
\end{equation}
\begin{equation}
    \operatorname{IVBS}_t(y_t^{\mathrm{g}})
    <
    r_{\mathrm{trig}}\mu_t.
\end{equation}
Here, $y_t^{\mathrm{g}}$ is the current greedy token, $\mathcal{H}_t$ stores the IVBS values of previously committed tokens, and $r_{\mathrm{trig}}$ controls triggering sensitivity. Upon triggering, we exclude the greedy token from the $k$ most probable candidates and discard candidates satisfying $\operatorname{IVBS}_t(y)<r_{\mathrm{rej}}\mu_t$, using IVBS as defined in Section~\ref{sec:ivbs}. The parameter $r_{\mathrm{rej}}$ controls the rejection threshold. If all candidates are rejected, we restore all non-greedy candidates.

From the retained set $\mathcal{C}_t^{+}$, we select a replacement by jointly considering language probability and binding score:
\begin{equation}
    y_t^\star
    =
    \arg\max_{y\in\mathcal{C}_t^{+}}
    \left[
    z\!\left(\log p_t(y)\right)
    +
    w\,z\!\left(\operatorname{IVBS}_t(y)\right)
    \right].
\end{equation}
Here, $p_t$ is the current prediction distribution, $z(\cdot)$ denotes standardization within the retained candidate set, and $w$ controls the weight of the binding score. BGR thus translates binding-state monitoring into local correction of the current token choice.

\subsection{Prefill Memory Reinjection}

Inspired by the markedly higher IVBS at first-token prediction than at subsequent steps, PMR treats prefill representations as a natural instruction--visual anchor and reintroduces them after rethinking corrects the current token choice, supplying instruction--visual information for subsequent predictions. Specifically, we cache the final hidden-state sequence after prefill as reference memory:
\begin{equation}
    \mathbf{M}
    =
    \mathbf{H}_{\mathrm{pre}}^{\mathrm{final}}
    \in\mathbb{R}^{N\times d}.
\end{equation}
Here, $N$ is the prefill sequence length and $d$ is the hidden dimension. These representations integrate IVSE's semantic enhancement from Section~\ref{sec:ivse} with processing by subsequent decoder layers. The memory is cached once per sample and remains fixed during subsequent generation.

PMR is triggered by token replacement in BGR. During the forward pass of the replacement token, memory is reinjected into the last-position hidden state after the final RMSNorm and before the output head:
\begin{equation}
    \mathbf{h}_t'
    =
    \operatorname{Mix}_{\rho}(\mathbf{h}_t,\mathbf{M}).
\end{equation}
Here, $\rho$ controls the reinjection strength, and $\operatorname{Mix}_{\rho}$ reuses the mixing operator defined in Section~\ref{sec:ivse}. This operation adjusts the next-token prediction distribution without rewriting existing KV cache entries.  \section{Experiments} 

\subsection{Experimental Setup}
\begin{table*}[t]
\centering
\caption{POPE evaluation on LLaVA-1.5 and InstructBLIP, reporting accuracy and F1 (\%) under Random, Popular, and Adversarial settings and their averages. Green shading marks the best results; arrows indicate absolute score changes from Regular.}
\label{tab:pope}
\fontsize{8.5}{9.5}\selectfont 
\renewcommand{\arraystretch}{1.02} 
\setlength{\tabcolsep}{2.0pt} 
\renewcommand{\ReBindDeltaSize}{5.2} 
\renewcommand{\ReBindDeltaLeading}{6.0} 
\arrayrulecolor{black} 
\begin{tabular*}{\textwidth}{@{\extracolsep{\fill}}ll*{8}{c}@{}}
\toprule
\multirow{2}{*}{Model} & \multirow{2}{*}{Method}
& \multicolumn{2}{c}{Random}
& \multicolumn{2}{c}{Popular}
& \multicolumn{2}{c}{Adversarial}
& \multicolumn{2}{c}{Average} \\
\cmidrule(lr){3-4}\cmidrule(lr){5-6}
\cmidrule(lr){7-8}\cmidrule(lr){9-10}
& & Acc$\uparrow$ & F1$\uparrow$
& Acc$\uparrow$ & F1$\uparrow$
& Acc$\uparrow$ & F1$\uparrow$
& Acc$\uparrow$ & F1$\uparrow$ \\
\midrule
\multirow{12}{*}{LLaVA-1.5} & Regular & 83.5\,\ReBindDelta{noimprovegray}{\uparrow}{0.0} & 82.3\,\ReBindDelta{noimprovegray}{\uparrow}{0.0} & 80.0\,\ReBindDelta{noimprovegray}{\uparrow}{0.0} & 79.3\,\ReBindDelta{noimprovegray}{\uparrow}{0.0} & 76.0\,\ReBindDelta{noimprovegray}{\uparrow}{0.0} & 76.3\,\ReBindDelta{noimprovegray}{\uparrow}{0.0} & 79.8\,\ReBindDelta{noimprovegray}{\uparrow}{0.0} & 79.3\,\ReBindDelta{noimprovegray}{\uparrow}{0.0} \\
 & VCD & 86.8\,\ReBindDelta{improvegray}{\uparrow}{3.3} & 86.8\,\ReBindDelta{improvegray}{\uparrow}{4.5} & 82.7\,\ReBindDelta{improvegray}{\uparrow}{2.7} & 83.4\,\ReBindDelta{improvegray}{\uparrow}{4.1} & 77.3\,\ReBindDelta{improvegray}{\uparrow}{1.3} & 79.3\,\ReBindDelta{improvegray}{\uparrow}{3.0} & 82.3\,\ReBindDelta{improvegray}{\uparrow}{2.5} & 83.2\,\ReBindDelta{improvegray}{\uparrow}{3.9} \\
 & ICD & 85.1\,\ReBindDelta{improvegray}{\uparrow}{1.6} & 84.2\,\ReBindDelta{improvegray}{\uparrow}{1.9} & 81.1\,\ReBindDelta{improvegray}{\uparrow}{1.1} & 80.7\,\ReBindDelta{improvegray}{\uparrow}{1.4} & 78.2\,\ReBindDelta{improvegray}{\uparrow}{2.2} & 78.6\,\ReBindDelta{improvegray}{\uparrow}{2.3} & 81.5\,\ReBindDelta{improvegray}{\uparrow}{1.7} & 81.2\,\ReBindDelta{improvegray}{\uparrow}{1.9} \\
 & DoLa & 84.8\,\ReBindDelta{improvegray}{\uparrow}{1.3} & 84.2\,\ReBindDelta{improvegray}{\uparrow}{1.9} & 79.8\,\ReBindDelta{noimprovegray}{\downarrow}{0.2} & 80.6\,\ReBindDelta{improvegray}{\uparrow}{1.3} & 76.3\,\ReBindDelta{improvegray}{\uparrow}{0.3} & 76.2\,\ReBindDelta{noimprovegray}{\downarrow}{0.1} & 80.3\,\ReBindDelta{improvegray}{\uparrow}{0.5} & 80.3\,\ReBindDelta{improvegray}{\uparrow}{1.0} \\
 & OPERA & 87.5\,\ReBindDelta{improvegray}{\uparrow}{4.0} & 86.5\,\ReBindDelta{improvegray}{\uparrow}{4.2} & 84.2\,\ReBindDelta{improvegray}{\uparrow}{4.2} & 83.5\,\ReBindDelta{improvegray}{\uparrow}{4.2} & 80.9\,\ReBindDelta{improvegray}{\uparrow}{4.9} & 80.7\,\ReBindDelta{improvegray}{\uparrow}{4.4} & 84.2\,\ReBindDelta{improvegray}{\uparrow}{4.4} & 83.6\,\ReBindDelta{improvegray}{\uparrow}{4.3} \\
 & ONLY & 87.5\,\ReBindDelta{improvegray}{\uparrow}{4.0} & 88.0\,\ReBindDelta{improvegray}{\uparrow}{5.7} & 79.7\,\ReBindDelta{noimprovegray}{\downarrow}{0.3} & 82.3\,\ReBindDelta{improvegray}{\uparrow}{3.0} & 72.4\,\ReBindDelta{noimprovegray}{\downarrow}{3.6} & 77.5\,\ReBindDelta{improvegray}{\uparrow}{1.2} & 79.9\,\ReBindDelta{improvegray}{\uparrow}{0.1} & 82.6\,\ReBindDelta{improvegray}{\uparrow}{3.3} \\
 & AGLA & 88.5\,\ReBindDelta{improvegray}{\uparrow}{5.0} & 87.7\,\ReBindDelta{improvegray}{\uparrow}{5.4} & 85.1\,\ReBindDelta{improvegray}{\uparrow}{5.1} & 84.7\,\ReBindDelta{improvegray}{\uparrow}{5.4} & 81.1\,\ReBindDelta{improvegray}{\uparrow}{5.1} & 81.4\,\ReBindDelta{improvegray}{\uparrow}{5.1} & 84.9\,\ReBindDelta{improvegray}{\uparrow}{5.1} & 84.6\,\ReBindDelta{improvegray}{\uparrow}{5.3} \\
 & MemVR & 89.7\,\ReBindDelta{improvegray}{\uparrow}{6.2} & 89.2\,\ReBindDelta{improvegray}{\uparrow}{6.9} & 86.4\,\ReBindDelta{improvegray}{\uparrow}{6.4} & 86.1\,\ReBindDelta{improvegray}{\uparrow}{6.8} & 82.3\,\ReBindDelta{improvegray}{\uparrow}{6.3} & 84.0\,\ReBindDelta{improvegray}{\uparrow}{7.7} & 86.1\,\ReBindDelta{improvegray}{\uparrow}{6.3} & 86.4\,\ReBindDelta{improvegray}{\uparrow}{7.1} \\
 & VISTA & 89.6\,\ReBindDelta{improvegray}{\uparrow}{6.1} & 89.0\,\ReBindDelta{improvegray}{\uparrow}{6.7} & 86.0\,\ReBindDelta{improvegray}{\uparrow}{6.0} & 85.9\,\ReBindDelta{improvegray}{\uparrow}{6.6} & 82.8\,\ReBindDelta{improvegray}{\uparrow}{6.8} & 82.8\,\ReBindDelta{improvegray}{\uparrow}{6.5} & 86.2\,\ReBindDelta{improvegray}{\uparrow}{6.4} & 85.9\,\ReBindDelta{improvegray}{\uparrow}{6.6} \\
 & LLaVA-Saliency & 87.5\,\ReBindDelta{improvegray}{\uparrow}{4.0} & 86.6\,\ReBindDelta{improvegray}{\uparrow}{4.3} & 86.0\,\ReBindDelta{improvegray}{\uparrow}{6.0} & 84.8\,\ReBindDelta{improvegray}{\uparrow}{5.5} & 83.4\,\ReBindDelta{improvegray}{\uparrow}{7.4} & 82.6\,\ReBindDelta{improvegray}{\uparrow}{6.3} & 85.6\,\ReBindDelta{improvegray}{\uparrow}{5.8} & 84.7\,\ReBindDelta{improvegray}{\uparrow}{5.4} \\
 & GACD & 87.4\,\ReBindDelta{improvegray}{\uparrow}{3.9} & 86.0\,\ReBindDelta{improvegray}{\uparrow}{3.7} & 86.2\,\ReBindDelta{improvegray}{\uparrow}{6.2} & 84.8\,\ReBindDelta{improvegray}{\uparrow}{5.5} & \ReBindBest{83.5}\,\ReBindDelta{improvegray}{\uparrow}{7.5} & 82.1\,\ReBindDelta{improvegray}{\uparrow}{5.8} & 85.7\,\ReBindDelta{improvegray}{\uparrow}{5.9} & 84.3\,\ReBindDelta{improvegray}{\uparrow}{5.0} \\
 & \textbf{Ours} & \ReBindBest{90.5}\,\ReBindDelta{improvegray}{\uparrow}{7.0} & \ReBindBest{90.6}\,\ReBindDelta{improvegray}{\uparrow}{8.3} & \ReBindBest{88.2}\,\ReBindDelta{improvegray}{\uparrow}{8.2} & \ReBindBest{88.2}\,\ReBindDelta{improvegray}{\uparrow}{8.9} & \ReBindBest{83.5}\,\ReBindDelta{improvegray}{\uparrow}{7.5} & \ReBindBest{84.2}\,\ReBindDelta{improvegray}{\uparrow}{7.9} & \ReBindBest{87.4}\,\ReBindDelta{improvegray}{\uparrow}{7.6} & \ReBindBest{87.7}\,\ReBindDelta{improvegray}{\uparrow}{8.4} \\
\midrule
\multirow{10}{*}{InstructBLIP} & Regular & 80.4\,\ReBindDelta{noimprovegray}{\uparrow}{0.0} & 80.9\,\ReBindDelta{noimprovegray}{\uparrow}{0.0} & 76.1\,\ReBindDelta{noimprovegray}{\uparrow}{0.0} & 77.7\,\ReBindDelta{noimprovegray}{\uparrow}{0.0} & 72.4\,\ReBindDelta{noimprovegray}{\uparrow}{0.0} & 75.4\,\ReBindDelta{noimprovegray}{\uparrow}{0.0} & 76.3\,\ReBindDelta{noimprovegray}{\uparrow}{0.0} & 78.0\,\ReBindDelta{noimprovegray}{\uparrow}{0.0} \\
 & VCD & 84.1\,\ReBindDelta{improvegray}{\uparrow}{3.7} & 84.1\,\ReBindDelta{improvegray}{\uparrow}{3.2} & 79.9\,\ReBindDelta{improvegray}{\uparrow}{3.8} & 80.8\,\ReBindDelta{improvegray}{\uparrow}{3.1} & 76.3\,\ReBindDelta{improvegray}{\uparrow}{3.9} & 78.1\,\ReBindDelta{improvegray}{\uparrow}{2.7} & 80.1\,\ReBindDelta{improvegray}{\uparrow}{3.8} & 81.0\,\ReBindDelta{improvegray}{\uparrow}{3.0} \\
 & ICD & 82.4\,\ReBindDelta{improvegray}{\uparrow}{2.0} & 83.3\,\ReBindDelta{improvegray}{\uparrow}{2.4} & 78.1\,\ReBindDelta{improvegray}{\uparrow}{2.0} & 79.1\,\ReBindDelta{improvegray}{\uparrow}{1.4} & 75.1\,\ReBindDelta{improvegray}{\uparrow}{2.7} & 76.2\,\ReBindDelta{improvegray}{\uparrow}{0.8} & 78.5\,\ReBindDelta{improvegray}{\uparrow}{2.2} & 79.6\,\ReBindDelta{improvegray}{\uparrow}{1.6} \\
 & DoLa & 83.0\,\ReBindDelta{improvegray}{\uparrow}{2.6} & 83.0\,\ReBindDelta{improvegray}{\uparrow}{2.1} & 79.0\,\ReBindDelta{improvegray}{\uparrow}{2.9} & 79.9\,\ReBindDelta{improvegray}{\uparrow}{2.2} & 74.7\,\ReBindDelta{improvegray}{\uparrow}{2.3} & 76.7\,\ReBindDelta{improvegray}{\uparrow}{1.3} & 78.9\,\ReBindDelta{improvegray}{\uparrow}{2.6} & 79.8\,\ReBindDelta{improvegray}{\uparrow}{1.8} \\
 & OPERA & 85.1\,\ReBindDelta{improvegray}{\uparrow}{4.7} & 84.4\,\ReBindDelta{improvegray}{\uparrow}{3.5} & 78.3\,\ReBindDelta{improvegray}{\uparrow}{2.2} & 80.2\,\ReBindDelta{improvegray}{\uparrow}{2.5} & 75.5\,\ReBindDelta{improvegray}{\uparrow}{3.1} & 78.2\,\ReBindDelta{improvegray}{\uparrow}{2.8} & 79.6\,\ReBindDelta{improvegray}{\uparrow}{3.3} & 80.9\,\ReBindDelta{improvegray}{\uparrow}{2.9} \\
 & ONLY & 87.2\,\ReBindDelta{improvegray}{\uparrow}{6.8} & 88.2\,\ReBindDelta{improvegray}{\uparrow}{7.3} & 80.4\,\ReBindDelta{improvegray}{\uparrow}{4.3} & 82.1\,\ReBindDelta{improvegray}{\uparrow}{4.4} & 74.8\,\ReBindDelta{improvegray}{\uparrow}{2.4} & 78.0\,\ReBindDelta{improvegray}{\uparrow}{2.6} & 80.8\,\ReBindDelta{improvegray}{\uparrow}{4.5} & 82.7\,\ReBindDelta{improvegray}{\uparrow}{4.7} \\
 & AGLA & 87.3\,\ReBindDelta{improvegray}{\uparrow}{6.9} & 87.1\,\ReBindDelta{improvegray}{\uparrow}{6.2} & 81.9\,\ReBindDelta{improvegray}{\uparrow}{5.8} & 82.6\,\ReBindDelta{improvegray}{\uparrow}{4.9} & 77.3\,\ReBindDelta{improvegray}{\uparrow}{4.9} & 79.2\,\ReBindDelta{improvegray}{\uparrow}{3.8} & 82.2\,\ReBindDelta{improvegray}{\uparrow}{5.9} & 82.9\,\ReBindDelta{improvegray}{\uparrow}{4.9} \\
 & VISTA & 87.5\,\ReBindDelta{improvegray}{\uparrow}{7.1} & 87.7\,\ReBindDelta{improvegray}{\uparrow}{6.8} & 81.3\,\ReBindDelta{improvegray}{\uparrow}{5.2} & 80.9\,\ReBindDelta{improvegray}{\uparrow}{3.2} & 80.8\,\ReBindDelta{improvegray}{\uparrow}{8.4} & 79.7\,\ReBindDelta{improvegray}{\uparrow}{4.3} & 83.2\,\ReBindDelta{improvegray}{\uparrow}{6.9} & 82.8\,\ReBindDelta{improvegray}{\uparrow}{4.8} \\
 & GACD & 87.9\,\ReBindDelta{improvegray}{\uparrow}{7.5} & 86.8\,\ReBindDelta{improvegray}{\uparrow}{5.9} & 85.0\,\ReBindDelta{improvegray}{\uparrow}{8.9} & 84.3\,\ReBindDelta{improvegray}{\uparrow}{6.6} & \ReBindBest{82.5}\,\ReBindDelta{improvegray}{\uparrow}{10.1} & 82.1\,\ReBindDelta{improvegray}{\uparrow}{6.7} & \ReBindBest{85.1}\,\ReBindDelta{improvegray}{\uparrow}{8.8} & 84.4\,\ReBindDelta{improvegray}{\uparrow}{6.4} \\
 & \textbf{Ours} & \ReBindBest{88.7}\,\ReBindDelta{improvegray}{\uparrow}{8.3} & \ReBindBest{88.8}\,\ReBindDelta{improvegray}{\uparrow}{7.9} & \ReBindBest{85.4}\,\ReBindDelta{improvegray}{\uparrow}{9.3} & \ReBindBest{85.6}\,\ReBindDelta{improvegray}{\uparrow}{7.9} & 81.3\,\ReBindDelta{improvegray}{\uparrow}{8.9} & \ReBindBest{82.3}\,\ReBindDelta{improvegray}{\uparrow}{6.9} & \ReBindBest{85.1}\,\ReBindDelta{improvegray}{\uparrow}{8.8} & \ReBindBest{85.6}\,\ReBindDelta{improvegray}{\uparrow}{7.6} \\
\bottomrule
\end{tabular*}
\vspace{-4pt}
\end{table*}
\paragraph{Model Architectures and Baselines.}
We evaluate ReBind on three representative large vision-language models with distinct architectural designs: InstructBLIP~\cite{dai2023instructblip}, which uses an instruction-aware Q-Former for cross-modal interaction; LLaVA-1.5~\cite{liu2024improved}, which employs a two-layer MLP projector for visual--textual alignment; and Qwen2.5-VL, which combines a native dynamic-resolution vision encoder with an MLP-based vision--language merger. To assess ReBind's effectiveness, we compare it with regular decoding and ten representative training-free hallucination mitigation methods, including contrastive decoding methods (VCD~\cite{leng2024mitigating}, ICD~\cite{wang2024mitigating}, and DoLa~\cite{chuang2024dola}); attention modulation and visual information enhancement methods (OPERA~\cite{huang2024opera}, ONLY~\cite{wan2025only}, AGLA~\cite{an2025mitigating}, MemVR~\cite{zou2024look}, and VISTA~\cite{li2025hidden}); and gradient-based methods (LLaVA-Saliency~\cite{zhang2026hallucination} and GACD~\cite{wang2026mitigating}).

\paragraph{Implementation Details.}
We evaluate ReBind on four widely used benchmarks: POPE~\cite{li2023evaluating}, CHAIR~\cite{rohrbach2018object}, and AMBER~\cite{wang2023amber} for hallucination evaluation, and MME~\cite{fu2026mme} for general multimodal capability assessment. Unless otherwise specified, IVBS is computed in the final three decoder layers of each model, and greedy decoding is used with a maximum generation length of 128 tokens. For IVSE, the semantic enhancement strength $\beta$ and uncertainty threshold $\tau$ are set to $0.20$ and $0.35$, respectively. For PMR, the reinjection ratio $\rho$ is set to $0.20$. For BGR, the trigger ratio $r_{\mathrm{trig}}$, candidate rejection ratio $r_{\mathrm{rej}}$, and binding score weight $w$ are set to $0.06$, $0.90$, and $0.15$, respectively.

\subsection{Results on Object Hallucination Benchmarks}

\paragraph{POPE Evaluation.}
POPE evaluates object hallucination through yes/no questions about object existence under three settings: Random, Popular, and Adversarial. As shown in Table~\ref{tab:pope}, our method outperforms regular decoding on both LLaVA-1.5 and InstructBLIP, improving accuracy and F1 by an average of 10.53\% and 10.17\%, respectively. Moreover, our method achieves the highest F1 across all three settings on both models, with average F1 scores of 87.7\% and 85.6\%, respectively, while attaining the best or joint-best average accuracy. These results demonstrate that our method effectively mitigates object hallucination across different model architectures and sampling settings.

\paragraph{CHAIR Evaluation.}
CHAIR measures object hallucination in open-ended image descriptions at the CHAIR sentence-level (\(C_s\)) and CHAIR instance-level (\(C_i\)), with lower values indicating better performance. As shown in Table~\ref{tab:chair_mme}, our method achieves the best results on both metrics for both models. Compared with regular decoding, the mean reductions in \(C_s\) and \(C_i\) across models reach 23.16\% and 35.33\%, respectively. Specifically, \(C_i\) decreases to 10.0\% on LLaVA-1.5 and 14.2\% on InstructBLIP, further validating the effectiveness of our method in suppressing object hallucination during open-ended generation across different model architectures.

\begin{table}[t]
\centering
\caption{CHAIR and MME evaluation on LLaVA-1.5 and InstructBLIP. $C_s$ and $C_i$ denote sentence- and instance-level hallucination rates. Green shading marks the best results; arrows indicate absolute score changes from Regular.}
\label{tab:chair_mme}

\fontsize{6.1}{7.2}\selectfont 
\renewcommand{\arraystretch}{1.04}
\setlength{\tabcolsep}{0.3pt} 
\renewcommand{\ReBindDeltaSize}{4.3} 
\renewcommand{\ReBindDeltaLeading}{5.1} 
\arrayrulecolor{black} 
\begin{tabular*}{\columnwidth}{@{\extracolsep{\fill}}lcccccc@{}}
\toprule
& \multicolumn{3}{c}{LLaVA-1.5}
& \multicolumn{3}{c}{InstructBLIP} \\
\cmidrule(lr){2-4}\cmidrule(lr){5-7}
Method
& \(C_s\downarrow\) & \(C_i\downarrow\) & MME$\uparrow$
& \(C_s\downarrow\) & \(C_i\downarrow\) & MME$\uparrow$ \\
\midrule
Regular & 51.0\,\ReBindDelta{deltazero}{\uparrow}{0.0} & 14.2\,\ReBindDelta{deltazero}{\uparrow}{0.0} & 1810.4\,\ReBindDelta{deltazero}{\uparrow}{0.0} & 54.2\,\ReBindDelta{deltazero}{\uparrow}{0.0} & 24.1\,\ReBindDelta{deltazero}{\uparrow}{0.0} & 1509.9\,\ReBindDelta{deltazero}{\uparrow}{0.0} \\
VCD & 56.6\,\ReBindDelta{deltaloss}{\uparrow}{5.6} & 15.2\,\ReBindDelta{deltaloss}{\uparrow}{1.0} & 1818.7\,\ReBindDelta{deltagain}{\uparrow}{8.3} & 53.4\,\ReBindDelta{deltagain}{\downarrow}{0.8} & 17.4\,\ReBindDelta{deltagain}{\downarrow}{6.7} & 1515.5\,\ReBindDelta{deltagain}{\uparrow}{5.6} \\
ICD & 51.8\,\ReBindDelta{deltaloss}{\uparrow}{0.8} & 13.8\,\ReBindDelta{deltagain}{\downarrow}{0.4} & 1757.3\,\ReBindDelta{deltaloss}{\downarrow}{53.1} & 72.2\,\ReBindDelta{deltaloss}{\uparrow}{18.0} & 22.8\,\ReBindDelta{deltagain}{\downarrow}{1.3} & -- \\
DoLa & 47.6\,\ReBindDelta{deltagain}{\downarrow}{3.4} & 13.4\,\ReBindDelta{deltagain}{\downarrow}{0.8} & 1829.7\,\ReBindDelta{deltagain}{\uparrow}{19.3} & 66.2\,\ReBindDelta{deltaloss}{\uparrow}{12.0} & 22.7\,\ReBindDelta{deltagain}{\downarrow}{1.4} & 1508.5\,\ReBindDelta{deltaloss}{\downarrow}{1.4} \\
OPERA & 45.6\,\ReBindDelta{deltagain}{\downarrow}{5.4} & 13.4\,\ReBindDelta{deltagain}{\downarrow}{0.8} & 1844.5\,\ReBindDelta{deltagain}{\uparrow}{34.1} & 49.2\,\ReBindDelta{deltagain}{\downarrow}{5.0} & 14.9\,\ReBindDelta{deltagain}{\downarrow}{9.2} & 1513.2\,\ReBindDelta{deltagain}{\uparrow}{3.3} \\
ONLY & 51.6\,\ReBindDelta{deltaloss}{\uparrow}{0.6} & 15.4\,\ReBindDelta{deltaloss}{\uparrow}{1.2} & 1876.4\,\ReBindDelta{deltagain}{\uparrow}{66.0} & 57.0\,\ReBindDelta{deltaloss}{\uparrow}{2.8} & 20.1\,\ReBindDelta{deltagain}{\downarrow}{4.0} & 1543.7\,\ReBindDelta{deltagain}{\uparrow}{33.8} \\
AGLA & 49.0\,\ReBindDelta{deltagain}{\downarrow}{2.0} & 14.2\,\ReBindDelta{deltazero}{\uparrow}{0.0} & 1864.7\,\ReBindDelta{deltagain}{\uparrow}{54.3} & 55.0\,\ReBindDelta{deltaloss}{\uparrow}{0.8} & 17.4\,\ReBindDelta{deltagain}{\downarrow}{6.7} & 1520.9\,\ReBindDelta{deltagain}{\uparrow}{11.0} \\
MemVR & 54.4\,\ReBindDelta{deltaloss}{\uparrow}{3.4} & 14.9\,\ReBindDelta{deltaloss}{\uparrow}{0.7} & 1857.6\,\ReBindDelta{deltagain}{\uparrow}{47.2} & -- & -- & -- \\
LLaVA-Saliency & 49.0\,\ReBindDelta{deltagain}{\downarrow}{2.0} & 14.2\,\ReBindDelta{deltazero}{\uparrow}{0.0} & 1797.2\,\ReBindDelta{deltaloss}{\downarrow}{13.2} & -- & -- & -- \\
GACD & 41.0\,\ReBindDelta{deltagain}{\downarrow}{10.0} & 10.9\,\ReBindDelta{deltagain}{\downarrow}{3.3} & -- & -- & -- & -- \\
\midrule
\textbf{Ours} & \ReBindBest{33.4}\,\ReBindDelta{deltagain}{\downarrow}{17.6} & \ReBindBest{10.0}\,\ReBindDelta{deltagain}{\downarrow}{4.2} & \ReBindBest{1897.3}\,\ReBindDelta{deltagain}{\uparrow}{86.9} & \ReBindBest{47.8}\,\ReBindDelta{deltagain}{\downarrow}{6.4} & \ReBindBest{14.2}\,\ReBindDelta{deltagain}{\downarrow}{9.9} & \ReBindBest{1555.6}\,\ReBindDelta{deltagain}{\uparrow}{45.7} \\
\bottomrule
\end{tabular*}
\vspace{-4pt}
\end{table}

\subsection{Results on General Multimodal Benchmarks}

To examine how hallucination mitigation affects general model capabilities, we further evaluate our method on MME, which covers 14 perception and cognition tasks. As shown in Table~\ref{tab:chair_mme}, our method achieves scores of 1897.3 and 1555.6 on LLaVA-1.5 and InstructBLIP, respectively, exceeding regular decoding by 86.9 and 45.7 points and outperforming all compared hallucination mitigation methods. Together with the POPE and CHAIR results, these findings indicate that our framework improves overall multimodal performance while mitigating object hallucination, demonstrating its broad applicability across models.

\subsection{Component Ablation}

To assess the contribution of each module, we progressively introduce IVSE, PMR, and BGR into LLaVA-1.5. As shown in Table~\ref{tab:component_ablation}, IVSE alone yields limited improvements. Adding PMR reduces \(C_s\) and \(C_i\) to 39.0\% and 9.8\%, respectively, indicating that reusing prefill memory effectively suppresses object hallucination. Further incorporating BGR lowers \(C_s\) by 14.36\% to 33.4\%, although \(C_i\) increases slightly to 10.0\%. The complete framework achieves the lowest sentence-level hallucination rate while maintaining a low instance-level hallucination rate, validating the combination of memory reinjection and binding-guided token correction.

\begin{table}[t]
\centering
\caption{Component ablation on LLaVA-1.5 using CHAIR (\%), evaluating the progressive inclusion of IVSE, PMR, and BGR. Best results are shown in bold.}
\label{tab:component_ablation}
\label{tab:component_ablation}
\fontsize{7.5}{8.5}\selectfont 
\renewcommand{\arraystretch}{0.85} 
\setlength{\tabcolsep}{2.0pt} 
\arrayrulecolor{black}
\begin{tabular*}{\columnwidth}{@{\extracolsep{\fill}}cccccc@{}}
\toprule
Baseline & IVSE & PMR & BGR
& \(C_s\downarrow\)
& \(C_i\downarrow\) \\
\midrule
$\checkmark$ & & &
& 51.0 & 14.2 \\
$\checkmark$ & $\checkmark$ & &
& 50.8 & 14.2 \\
$\checkmark$ & $\checkmark$ & $\checkmark$ &
& 39.0 & \textbf{9.8} \\
$\checkmark$ & $\checkmark$ & $\checkmark$ & $\checkmark$
& \textbf{33.4} & 10.0 \\
\bottomrule
\end{tabular*}
\vspace{-4pt}
\end{table}

\subsection{Computational Overhead}

\begin{figure}[t]
    \centering
    \includegraphics[width=\columnwidth]{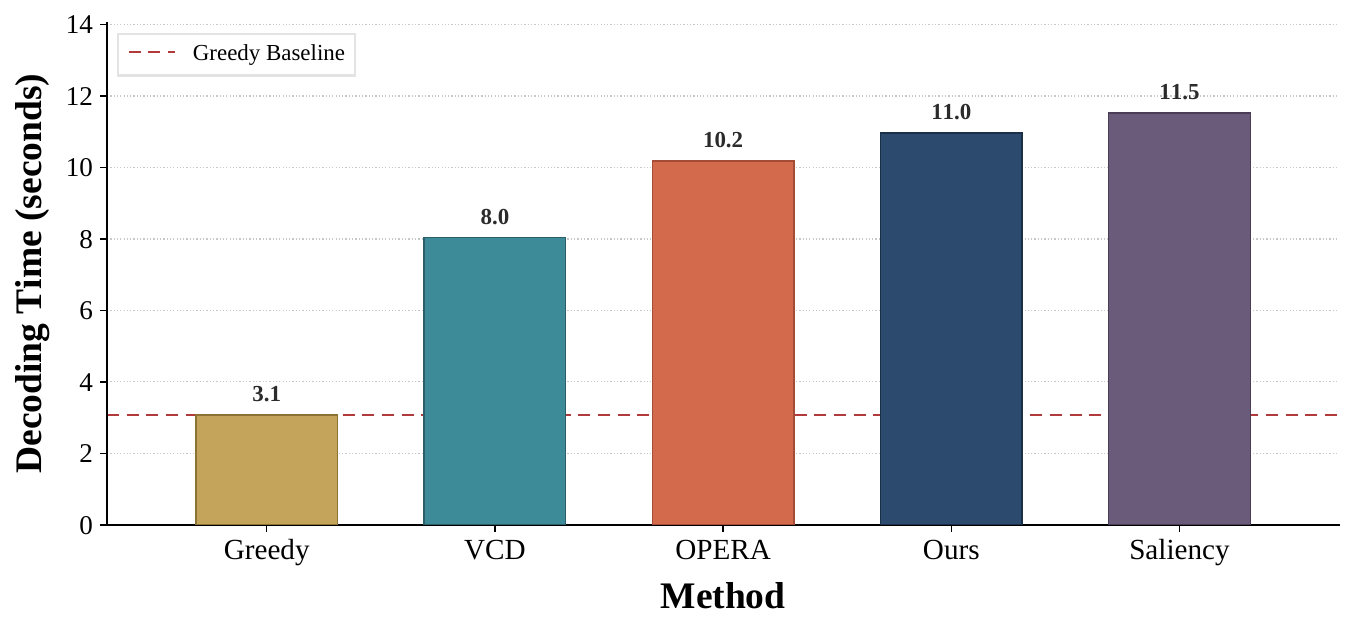}
    \caption{Average generation time per sample on CHAIR. The dashed line marks the greedy-decoding baseline; lower is better.}
\label{fig:eff_time1}
    \vspace{-8pt}
\end{figure}

To analyze the impact of different hallucination mitigation methods on LVLM inference efficiency, we measure the average generation time per sample on the CHAIR generation task. As shown in Fig.~\ref{fig:eff_time1}, our method takes longer than VCD, a contrastive decoding method, and OPERA, an attention-based intervention method, but achieves better hallucination mitigation. Compared with the similarly gradient-based LLaVA-Saliency, our method reduces average generation time by approximately 4.3\% while outperforming it on all reported hallucination metrics. These results indicate that, despite the additional inference overhead introduced by local gradient computation, our method achieves a favorable trade-off between hallucination suppression and generation efficiency.

\subsection{Scalability Analysis}

To examine the framework's applicability across model architectures and parameter scales, we extend our evaluation to Qwen2.5-VL-7B and Qwen2.5-VL-3B. As shown in Table~\ref{tab:qwen_generalization}, our method achieves a POPE F1 of 86.7\% on the 7B model and reduces \(C_s\) and \(C_i\) to 20.8\% and 7.4\%, respectively, outperforming LLaVA-Saliency on all three metrics. On the smaller 3B model, it improves POPE F1 by 0.69\% over regular decoding and reduces the two CHAIR metrics by 8.47\% and 9.72\%, respectively. In addition, MME scores increase by 19.9 and 8.7 points on the 7B and 3B models. These results demonstrate consistent performance gains across architectures and model scales, improving overall multimodal performance while mitigating object hallucination during text generation.

\begin{table}[t]
\centering
\caption{POPE, CHAIR, and MME evaluation on Qwen2.5-VL-7B and Qwen2.5-VL-3B. Green shading marks the best results; arrows indicate absolute score changes from Regular.}
\label{tab:qwen_generalization}
\fontsize{6.0}{7.0}\selectfont 
\renewcommand{\arraystretch}{1.04} 
\setlength{\tabcolsep}{0.6pt} 
\renewcommand{\ReBindDeltaSize}{4.2} 
\renewcommand{\ReBindDeltaLeading}{5.1} 
\arrayrulecolor{black}
\begin{tabular*}{\columnwidth}{@{\extracolsep{\fill}}llccccc@{}}
\toprule
\multirow{2}{*}{Model} & \multirow{2}{*}{Method}
& \multicolumn{2}{c}{POPE}
& \multicolumn{2}{c}{CHAIR}
& MME \\
\cmidrule(lr){3-4}\cmidrule(lr){5-6}\cmidrule(lr){7-7}
& & Acc$\uparrow$ & F1$\uparrow$
& \(C_s\downarrow\) & \(C_i\downarrow\) & Total$\uparrow$ \\
\midrule
\multirow{3}{*}{\shortstack[l]{Qwen2.5-VL\\7B}} & Regular & 86.1\,\ReBindDelta{deltazero}{\uparrow}{0.0} & 84.7\,\ReBindDelta{deltazero}{\uparrow}{0.0} & 27.4\,\ReBindDelta{deltazero}{\uparrow}{0.0} & 8.7\,\ReBindDelta{deltazero}{\uparrow}{0.0} & 2309.4\,\ReBindDelta{deltazero}{\uparrow}{0.0} \\
 & LLaVA-Saliency & 86.2\,\ReBindDelta{deltagain}{\uparrow}{0.1} & 84.9\,\ReBindDelta{deltagain}{\uparrow}{0.2} & 23.8\,\ReBindDelta{deltagain}{\downarrow}{3.6} & 7.7\,\ReBindDelta{deltagain}{\downarrow}{1.0} & -- \\
 & \textbf{Ours} & \ReBindBest{87.8}\,\ReBindDelta{deltagain}{\uparrow}{1.7} & \ReBindBest{86.7}\,\ReBindDelta{deltagain}{\uparrow}{2.0} & \ReBindBest{20.8}\,\ReBindDelta{deltagain}{\downarrow}{6.6} & \ReBindBest{7.4}\,\ReBindDelta{deltagain}{\downarrow}{1.3} & \ReBindBest{2329.3}\,\ReBindDelta{deltagain}{\uparrow}{19.9} \\
\midrule
\multirow{3}{*}{\shortstack[l]{Qwen2.5-VL\\3B}} & Regular & 85.5\,\ReBindDelta{deltazero}{\uparrow}{0.0} & 86.9\,\ReBindDelta{deltazero}{\uparrow}{0.0} & 23.6\,\ReBindDelta{deltazero}{\uparrow}{0.0} & 7.2\,\ReBindDelta{deltazero}{\uparrow}{0.0} & 1591.7\,\ReBindDelta{deltazero}{\uparrow}{0.0} \\
 & LLaVA-Saliency & 82.5\,\ReBindDelta{deltaloss}{\downarrow}{3.0} & 80.8\,\ReBindDelta{deltaloss}{\downarrow}{6.1} & 25.4\,\ReBindDelta{deltaloss}{\uparrow}{1.8} & 7.8\,\ReBindDelta{deltaloss}{\uparrow}{0.6} & -- \\
 & \textbf{Ours} & \ReBindBest{86.4}\,\ReBindDelta{deltagain}{\uparrow}{0.9} & \ReBindBest{87.5}\,\ReBindDelta{deltagain}{\uparrow}{0.6} & \ReBindBest{21.6}\,\ReBindDelta{deltagain}{\downarrow}{2.0} & \ReBindBest{6.5}\,\ReBindDelta{deltagain}{\downarrow}{0.7} & \ReBindBest{1600.4}\,\ReBindDelta{deltagain}{\uparrow}{8.7} \\
\bottomrule
\end{tabular*}
\vspace{-4pt}
\end{table}

\begin{figure*}[t]
    \centering
    \begin{minipage}[t]{0.48\textwidth}
        \centering
        \includegraphics[width=\linewidth]{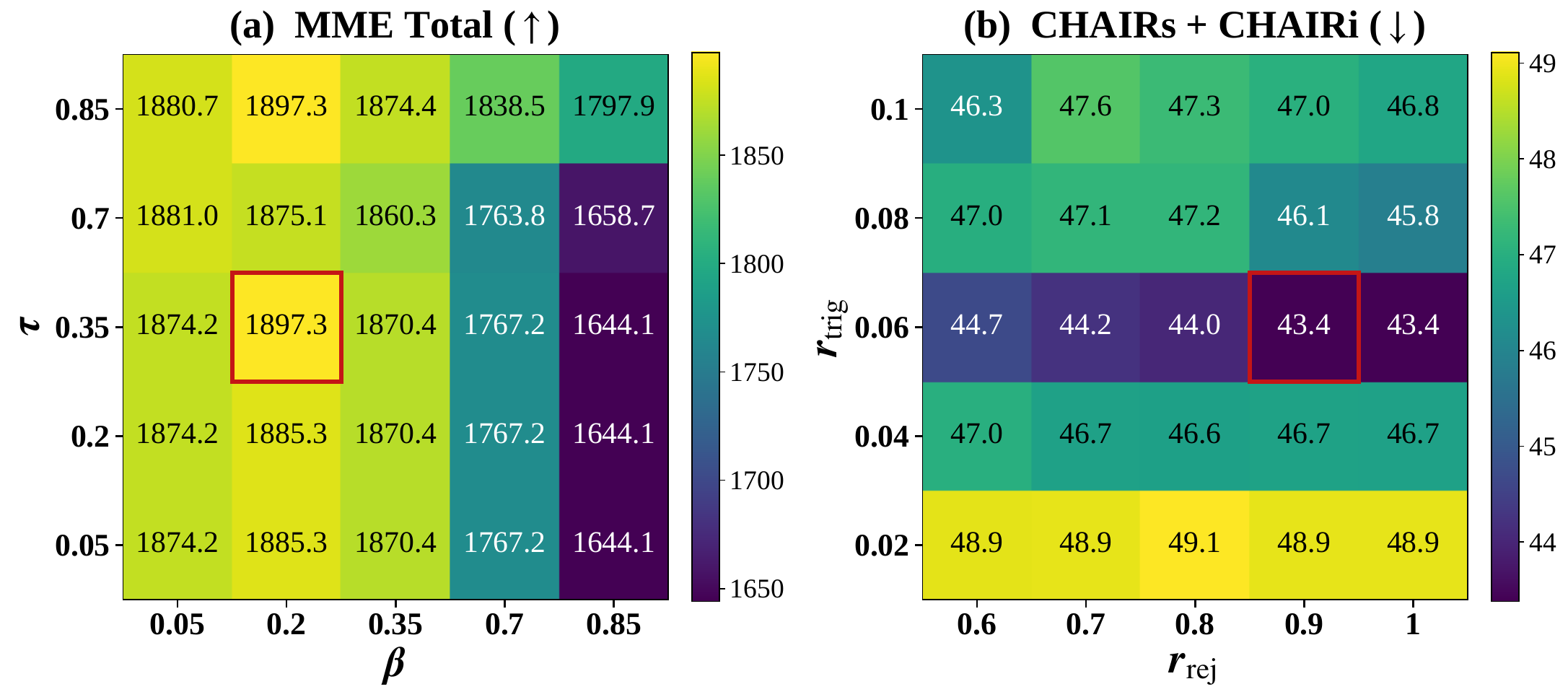}
       \caption{Hyperparameter analysis on LLaVA-1.5. Left: MME scores across $\beta\times\tau$. Right: combined CHAIR scores ($C_s+C_i$) across $r_{\mathrm{trig}}\times r_{\mathrm{rej}}$. Red boxes mark default settings.}
\label{fig:heatmaps}
    \end{minipage}
    \hfill
    \begin{minipage}[t]{0.48\textwidth}
        \centering
        \includegraphics[width=\linewidth]{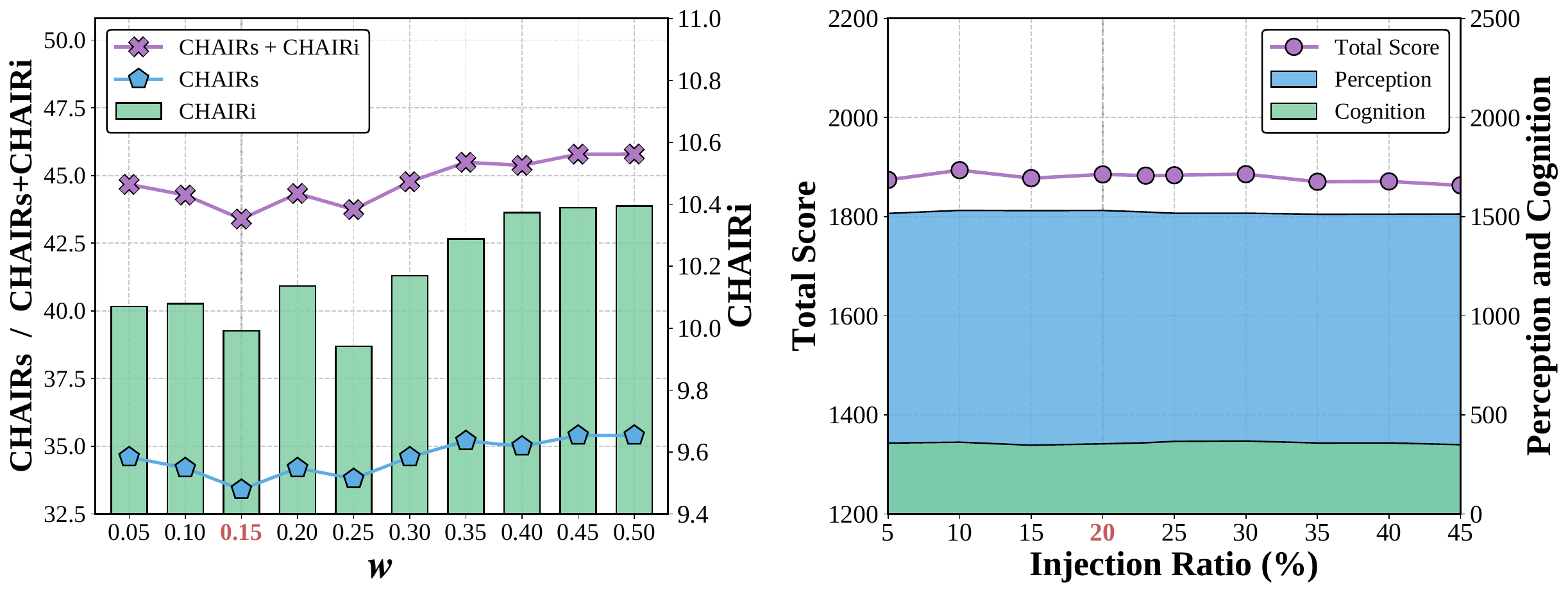}
       \caption{Sensitivity to $w$ and $\rho$ on LLaVA-1.5. Left: sentence-level, instance-level, and combined CHAIR scores across $w$. Right: MME total, perception, and cognition scores across $\rho$.}
\label{fig:curves}
    \end{minipage}
    \vspace{-8pt}
\end{figure*}

\subsection{Generalization and Qualitative Analysis}

To further validate ReBind's generalization across hallucination benchmarks, we extend our evaluation to AMBER, which assesses hallucination robustness across diverse scenarios. As shown in Table~\ref{tab:amber}, our method reduces CHAIR, Hal, and Cog on both LLaVA-1.5 and InstructBLIP while slightly improving Cover. Specifically, CHAIR decreases by 24.66\% and 60.00\% on the two models, respectively, while Cover increases from 50.9 to 51.8 and from 53.5 to 53.6. These results demonstrate that our method substantially reduces hallucinated objects while improving ground-truth object coverage, further validating its applicability across benchmarks.

\begin{table}[t]
\centering
\caption{AMBER generative evaluation on LLaVA-1.5 and InstructBLIP. Green shading marks the best results; arrows indicate absolute score changes from Regular.}
\label{tab:amber}
\fontsize{7.2}{8.3}\selectfont 
\renewcommand{\arraystretch}{1.02} 
\setlength{\tabcolsep}{1.0pt} 
\renewcommand{\ReBindDeltaSize}{4.8} 
\renewcommand{\ReBindDeltaLeading}{5.5} 
\arrayrulecolor{black}
\begin{tabular*}{\columnwidth}{@{\extracolsep{\fill}}llcccc@{}}
\toprule
\multirow{2}{*}{Model} & \multirow{2}{*}{Method}
& \multicolumn{4}{c}{AMBER} \\
\cmidrule(lr){3-6}
& & CHAIR$\downarrow$ & Cover$\uparrow$ & Hal$\downarrow$ & Cog$\downarrow$ \\
\midrule
\multirow{2}{*}{LLaVA-1.5} & Regular & 7.3\,\ReBindDelta{deltazero}{\uparrow}{0.0} & 50.9\,\ReBindDelta{deltazero}{\uparrow}{0.0} & 33.2\,\ReBindDelta{deltazero}{\uparrow}{0.0} & 3.9\,\ReBindDelta{deltazero}{\uparrow}{0.0} \\
 & \textbf{Ours} & \ReBindBest{5.5}\,\ReBindDelta{deltagain}{\downarrow}{1.8} & \ReBindBest{51.8}\,\ReBindDelta{deltagain}{\uparrow}{0.9} & \ReBindBest{30.0}\,\ReBindDelta{deltagain}{\downarrow}{3.2} & \ReBindBest{2.4}\,\ReBindDelta{deltagain}{\downarrow}{1.5} \\
\midrule
\multirow{2}{*}{InstructBLIP} & Regular & 18.5\,\ReBindDelta{deltazero}{\uparrow}{0.0} & 53.5\,\ReBindDelta{deltazero}{\uparrow}{0.0} & 57.7\,\ReBindDelta{deltazero}{\uparrow}{0.0} & 6.5\,\ReBindDelta{deltazero}{\uparrow}{0.0} \\
 & \textbf{Ours} & \ReBindBest{7.4}\,\ReBindDelta{deltagain}{\downarrow}{11.1} & \ReBindBest{53.6}\,\ReBindDelta{deltagain}{\uparrow}{0.1} & \ReBindBest{35.5}\,\ReBindDelta{deltagain}{\downarrow}{22.2} & \ReBindBest{3.8}\,\ReBindDelta{deltagain}{\downarrow}{2.7} \\
\bottomrule
\end{tabular*}
\vspace{-4pt}
\end{table}

Our method intervenes in generation through hidden-representation fusion and token replacement, which may affect the model's original language generation behavior and alter description length and completeness. To assess this effect, we compare the average word and token counts of descriptions generated by ReBind and the base model on AMBER. As shown in Table~\ref{tab:generation_length}, applying ReBind changes the average word count from 90.8 to 90.0 and the average token count from 117.2 to 116.0, both remaining close to the baseline. Together with the improvements in Cover, these results indicate that ReBind mitigates hallucination while maintaining a similar description length, without achieving its gains at the expense of substantially shorter outputs or reduced ground-truth object coverage.

\begin{table}[t]
\centering
\caption{Average word and token counts of descriptions generated by Regular decoding and ReBind on AMBER. ReBind maintains a description length comparable to the baseline.}
\label{tab:generation_length}
\fontsize{8.5}{9.5}\selectfont 
\renewcommand{\arraystretch}{0.92} 
\setlength{\tabcolsep}{2.0pt} 
\arrayrulecolor{black}
\begin{tabular*}{\columnwidth}{@{\extracolsep{\fill}}lcc@{}}
\toprule
Method & Average words & Average tokens \\
\midrule
Regular & 90.8 & 117.2 \\
Ours    & 90.0 & 116.0 \\
\bottomrule
\end{tabular*}
\vspace{-4pt}
\end{table}

\subsection{Discussion}

\textbf{Hyperparameter Analysis}. To assess the impact of hyperparameter settings on model performance, we examine two parameter combinations, $\beta\times\tau$ and $r_{\mathrm{trig}}\times r_{\mathrm{rej}}$, along with the score weight $w$ and reinjection ratio $\rho$ on LLaVA-1.5. As shown in Fig.~\ref{fig:heatmaps}, low-to-moderate values of $\beta$ maintain high MME scores, whereas overly strong semantic enhancement causes substantial degradation; adjusting $\tau$ has a limited effect when $\beta$ is small. Hallucination metrics vary non-monotonically with $r_{\mathrm{trig}}$, with better results at $0.06$, indicating that indiscriminately increasing rethinking triggers does not always improve performance. The effect of $r_{\mathrm{rej}}$ is relatively small. Moreover, Fig.~\ref{fig:curves} shows that $w=0.15$ achieves the lowest combined CHAIR score, and further increases do not yield consistent gains, while MME scores remain broadly stable across different values of $\rho$. These results indicate that semantic enhancement strength and the rethinking trigger ratio require appropriate tuning, but ReBind remains robust to variations in most hyperparameters despite their number.

\begin{figure}[t]
    \centering
    \includegraphics[width=\columnwidth]{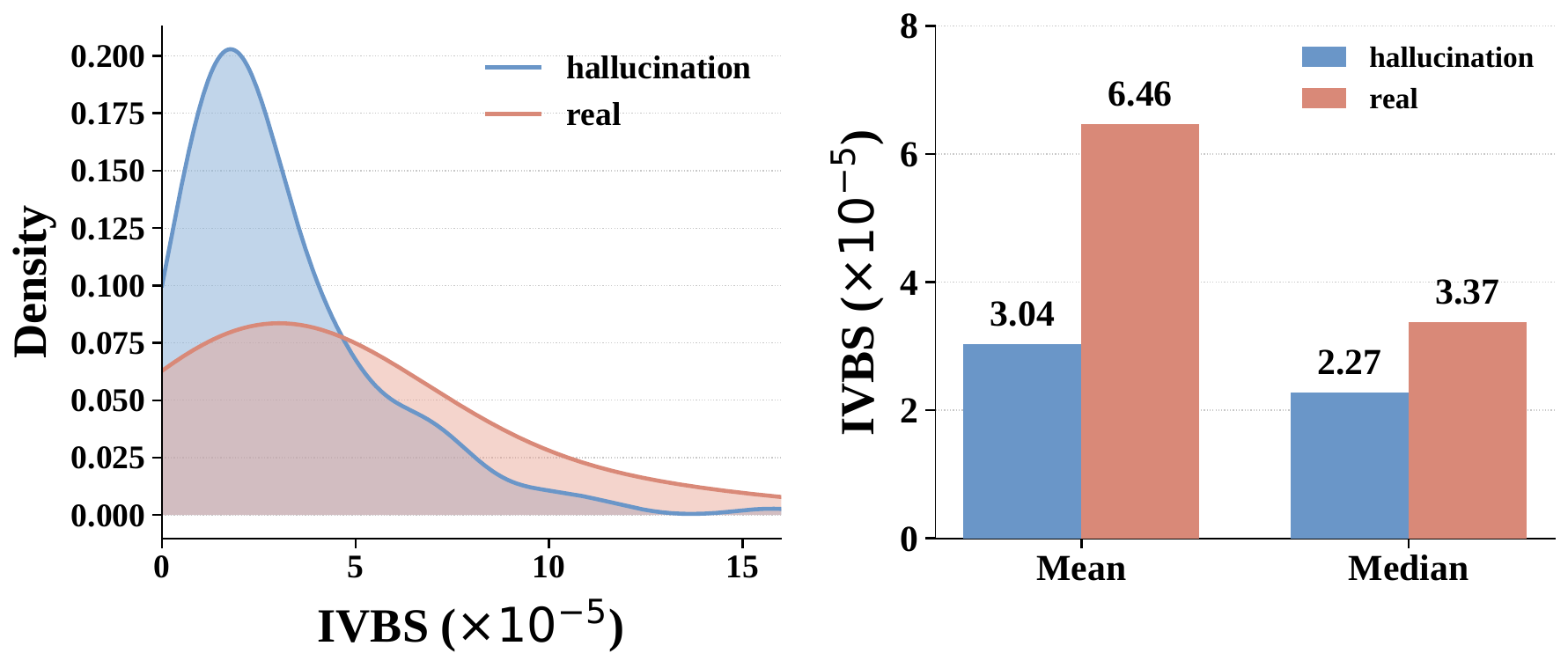}
    \caption{IVBS distributions (left) and means and medians (right) across 200 MSCOCO images. Hallucinated object tokens generally score lower than faithful ones.}
\label{fig:ivbs_overlap}
    \vspace{-8pt} 
\end{figure}

\noindent \textbf{Correlation Between IVBS and Hallucination} To further examine the association between IVBS and object hallucination, we sample 200 images from MSCOCO~\cite{lin2014microsoft}, generate descriptions, and analyze the IVBS distributions of faithful and hallucinated object tokens. As shown in Fig.~\ref{fig:ivbs_overlap}, hallucinated object tokens are more concentrated in the low-score region, whereas faithful object tokens exhibit higher central values. Specifically, the mean and median for hallucinated object tokens are $3.04\times10^{-5}$ and $2.27\times10^{-5}$, respectively, both below the corresponding values of $6.46\times10^{-5}$ and $3.37\times10^{-5}$ for faithful object tokens. This distributional difference indicates an association between weaker instruction--visual binding and hallucination generation, providing empirical support for using IVBS to monitor hallucination risk and guide decoding interventions during autoregressive generation.

\section{Conclusion}

In this work, we revisit LVLM hallucinations via visual-differential gradients and introduce IVBS to characterize instruction--visual binding, uncovering its weakening as a key driver of hallucinations. We propose ReBind, a training-free, two-stage framework. ReBind strengthens prefill representations via IVSE and combines BGR token correction with PMR memory reinjection during decoding to boost visual grounding. Specifically, BGR corrects token choices with weak binding responses, while PMR reuses prefill memory to support subsequent predictions. Experiments show that ReBind both mitigates hallucinations and preserves object coverage.

{
    \small
    \bibliographystyle{ieeenat_fullname}
    \bibliography{main}
}
\clearpage
\section{Implementation Details}
\subsection{Datasets}
\label{sec:supp_datasets}

\noindent \textbf{POPE.}
Polling-based Object Probing Evaluation (POPE) measures object hallucination
through yes/no questions about whether a specified object appears in an image.
We use the standard MS COCO evaluation split of 500 images. Under each setting,
every image is paired with six questions: three about objects present in the
image and three about absent objects, yielding 3,000 questions per setting.
The three settings differ in how absent objects are selected. \textit{Random}
samples them at random, \textit{Popular} selects those most frequent in the
dataset, and \textit{Adversarial} selects those that most frequently co-occur
with objects present in the image. We report accuracy and F1 for each setting,
along with the mean of each metric across the three settings.

\noindent \textbf{CHAIR.}
Caption Hallucination Assessment with Image Relevance (CHAIR) evaluates object
hallucination in open-ended image descriptions. The evaluator maps object
mentions to the 80 COCO categories. A mention is counted as hallucinated if
its category appears in neither the image's instance annotations nor its
human-written reference captions. CHAIR comprises two complementary metrics,
$\mathrm{CHAIR}_{I}$ and $\mathrm{CHAIR}_{S}$, denoted by $C_i$ and $C_s$ in
the main paper:
\begin{equation}
    \mathrm{CHAIR}_{I}
    =
    \frac{
        \left|\left\{\text{hallucinated objects}\right\}\right|
    }{
        \left|\left\{\text{all objects mentioned}\right\}\right|
    }.
\end{equation}
\begin{equation}
    \mathrm{CHAIR}_{S}
    =
    \frac{
        \left|\left\{\text{sentences with hallucinated objects}\right\}\right|
    }{
        \left|\left\{\text{all sentences}\right\}\right|
    }.
\end{equation}
$\mathrm{CHAIR}_{I}$ measures the fraction of object mentions that are
hallucinated, whereas $\mathrm{CHAIR}_{S}$ measures the fraction of captions
containing at least one hallucinated object. Lower values are better for both
metrics. We randomly sample 500 images from the MS COCO 2014 validation split
and prompt each model with \textit{``Please describe this image in detail.''}
We use greedy decoding and generate at most 128 tokens per image.

\noindent \textbf{MME.}
We use MME to assess how hallucination mitigation affects general multimodal
capabilities. The benchmark comprises 10 perception subtasks and four cognition
subtasks. Perception tasks include recognizing object existence, count,
position, and color, as well as optical character recognition. The cognition
tasks cover commonsense reasoning, numerical calculation, text translation,
and code reasoning. Each image is paired with two yes/no questions. Each
subtask score is the sum of question-level accuracy and paired accuracy; a
pair is correct only if both answers are correct. We report the total across
all 14 subtasks, with a maximum of 2,800 points: 2,000 for perception and 800
for cognition.

\noindent \textbf{AMBER.}
AMBER evaluates hallucinations without an external LLM judge. We use its
generative task, which contains 1,004 images with one description prompt per
image. Each image has annotations for both visible objects and objects that
are plausible in the scene but absent from the image. We report four metrics.
\textit{CHAIR} measures the fraction of recognized object mentions that are
hallucinated, while \textit{Cover} measures the fraction of annotated visible
objects mentioned in the descriptions. \textit{Hal} is the fraction of
descriptions containing at least one hallucinated object. \textit{Cog} measures
the fraction of annotated plausible but absent objects that the model
incorrectly mentions. Higher Cover and lower CHAIR, Hal, and Cog indicate
better performance. We also report the mean word and token counts per
description to examine whether hallucination reductions coincide with shorter
outputs.

\section{Additional Experiments}
\subsection{Ablation on IVBS Aggregation Layers}
To assess how layer selection affects hallucination mitigation and inference efficiency, we compute IVBS over the last 3, 5, 7, 9, or 11 decoder layers of LLaVA-1.5. As shown in Table~\ref{tab:ivbs_layers}, using the last three layers yields the lowest sentence-level hallucination rate ($\mathrm{CHAIR}_{S}=33.4\%$) and the fastest inference (11.0 seconds per sample, or 10.6 tokens per second). Extending the aggregation to the last five or seven layers increases both sentence- and instance-level hallucination rates. Using the last 11 layers reduces $\mathrm{CHAIR}_{I}$ from 10.0\% to its lowest value of 9.6\%, but raises $\mathrm{CHAIR}_{S}$ to 37.0\% and increases inference time to 13.6 seconds per sample. We therefore use the last three layers by default, achieving the best sentence-level performance and inference efficiency while maintaining a low instance-level hallucination rate.

\begin{table}[t]
    \centering
    \caption{Ablation on the number of decoder layers used to compute IVBS
    on LLaVA-1.5. CHAIR scores are percentages. Best results are in bold.}
    \label{tab:ivbs_layers}
    \fontsize{8}{9.2}\selectfont           
    \renewcommand{\arraystretch}{1.00}    
    \setlength{\tabcolsep}{2pt}           
    \begin{tabular*}{\columnwidth}{@{\extracolsep{\fill}}lcccc@{}}
        \toprule
        Layers & $\mathrm{CHAIR}_{S}\!\downarrow$
               & $\mathrm{CHAIR}_{I}\!\downarrow$
               & s/sample$\downarrow$ & tokens/s$\uparrow$ \\
        \midrule
        \textbf{Last 3 (ours)} & \textbf{33.4} & 10.0
                              & \textbf{11.0} & \textbf{10.6} \\
        Last 5  & 38.8 & 11.0         & 11.8 & 9.9 \\
        Last 7  & 38.8 & 11.4         & 12.4 & 9.4 \\
        Last 9  & 34.6 & 9.7          & 13.9 & 8.4 \\
        Last 11 & 37.0 & \textbf{9.6} & 13.6 & 8.6 \\
        \bottomrule
    \end{tabular*}
\end{table}

\subsection{Discussion}
\begin{figure}[t]
    \centering
    \includegraphics[width=\columnwidth]{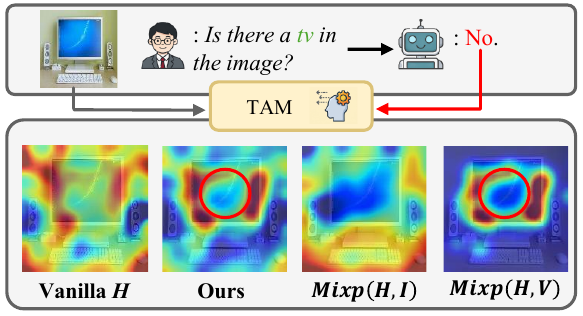}
    \caption{Token activation maps for the original representations, visual-only enhancement, instruction-only enhancement, and full IVSE.}
\label{fig:eff_time}
    \vspace{-8pt} 
\end{figure}

\noindent \textbf{Why IVSE works.}
To examine how visual and instruction representations complement each other in IVSE, we use Token Activation Maps (TAM) to compare four settings: the original representations, enhancement with the visual subset alone, enhancement with the instruction subset alone, and full IVSE. As shown in Fig.~\ref{fig:eff_time}, when the question refers to the ``tv,'' the original activations are diffuse: the target screen does not stand out, and the keyboard and background also show clear responses. With the visual subset alone, activations concentrate on the screen and its outline. With the instruction subset alone, they cover a broader region of the scene but are less focused on the screen. Combining both subsets makes the screen outline clearer than in the original representations and reduces responses in non-target regions such as the keyboard, while retaining visible activations around the screen. This comparison suggests that IVSE highlights the object specified in the question while preserving scene context, providing more relevant visual evidence for subsequent predictions.

\section{The Algorithm of ReBind}
\label{sec:supp_rebind_algorithm}

We present ReBind in two algorithms. Algorithm~\ref{alg:ivbs} computes the Instruction--Visual Binding Score (IVBS) for a candidate token using visual-differential gradients. Algorithm~\ref{alg:rebind} describes the full inference procedure. During prefill, IVSE strengthens instruction--visual representations, and the final representations are cached as reference memory. During decoding, BGR uses IVBS to identify and replace weakly bound tokens, after which PMR reinjects the cached memory to support the next prediction.

\begin{algorithm}[t]
\caption{Instruction--Visual Binding Score (IVBS)}
\label{alg:ivbs}
\small
\begin{algorithmic}[1]
\Require Current context $\mathbf{x}_t$, candidate token $y$,
         instruction positions $\mathcal{I}$, scoring layers
         $\mathcal{L}_{\mathrm{score}}$
\Ensure Binding score $b_t(y)$

\For{$\alpha \in \{1,0\}$}
    \State Retain ($\alpha=1$) or zero ($\alpha=0$) the visual
           keys and values in $\mathcal{L}_{\mathrm{score}}$,
           keeping all token positions fixed.
    \State Apply differentiable gates
           $\boldsymbol{\theta}^{\ell,h}=\mathbf{1}$ to the
           last attention row of each scoring head.
    \State Compute $\mathcal{L}^{\alpha}=-\log p_t^\alpha(y)$
           for the same candidate $y$.
    \State Compute $\mathbf{G}^{\alpha,\ell,h}
           \gets \partial\mathcal{L}^{\alpha}/
           \partial\boldsymbol{\theta}^{\ell,h}$
           by local backpropagation.
\EndFor

\For{$\ell \in \mathcal{L}_{\mathrm{score}}$}
    \State $\Delta\mathbf{G}^{\ell,h}
           \gets \mathbf{G}^{1,\ell,h}-\mathbf{G}^{0,\ell,h}$
           for each head $h$.
    \State $\mathbf{s}^{\ell}
           \gets \frac{1}{H}\sum_{h=1}^{H}
           \left|\mathbf{a}^{1,\ell,h}
           \odot\Delta\mathbf{G}^{\ell,h}\right|$
    \State $\bar{\mathbf{s}}^{\ell}
           \gets \mathbf{s}^{\ell}/
           \max(\|\mathbf{s}^{\ell}\|_2,\epsilon)$
    \State $b_{\ell}
           \gets \frac{1}{|\mathcal{I}|}
           \sum_{j\in\mathcal{I}}\bar{s}^{\ell}_{j}$
\EndFor
\State \Return $b_t(y)=\min_{\ell\in\mathcal{L}_{\mathrm{score}}}b_\ell$
\end{algorithmic}
\end{algorithm}

\begin{algorithm*}[!tp]
\caption{Two-Stage Inference with ReBind}
\label{alg:rebind}
\small 
\centering
\begin{minipage}{0.98\linewidth}
\setlength{\parindent}{0pt}
\setlength{\parskip}{2pt}              
\setlength{\abovedisplayskip}{3pt}     
\setlength{\belowdisplayskip}{3pt}     
\setlength{\abovedisplayshortskip}{2pt}
\setlength{\belowdisplayshortskip}{2pt}

\textbf{Input:} Image and instruction, frozen LVLM, maximum generation
length $T$, candidate count $k$, and parameters
$\beta,\tau,\rho,r_{\mathrm{trig}},r_{\mathrm{rej}},w$.\par
\textbf{Output:} Generated token sequence $Y$.\par

\medskip
\textbf{Stage 1: Prefill and semantic enhancement (IVSE)}\par

\textbf{1. Estimate prediction uncertainty.}
At each decoder layer $\ell$, project the last-position representation
to logits $\mathbf{z}^{(\ell)}$ and compute the normalized entropy
over the top 10 candidates:
\[
    \mathbf{q}^{(\ell)}
    = \operatorname{softmax}\!\left(
        \operatorname{TopK}(\mathbf{z}^{(\ell)},10)\right),
    \qquad
    U^{(\ell)}
    = -\frac{1}{\log 10}
      \sum_{i=1}^{10} q_i^{(\ell)}\log q_i^{(\ell)}.
\]

\textbf{2. Enhance instruction--visual representations.}
If $U^{(\ell)}>\tau$, concatenate the representations at visual
positions $\mathcal{V}$ and instruction positions $\mathcal{I}$,
then mix them into the full hidden sequence:
\[
    \mathbf{S}^{(\ell)}
    = [\mathbf{H}_{\mathcal{V}}^{(\ell)};
       \mathbf{H}_{\mathcal{I}}^{(\ell)}],
    \qquad
    \mathbf{H}^{(\ell)}
    \gets \operatorname{Mix}_{\beta}
          (\mathbf{H}^{(\ell)},\mathbf{S}^{(\ell)}).
\]
Otherwise, retain $\mathbf{H}^{(\ell)}$ unchanged.
Continue through the remaining decoder layers.\par

\textbf{3. Cache prefill memory.}
Store the final prefill sequence as fixed reference memory
$\mathbf{M}\gets\mathbf{H}_{\mathrm{pre}}^{\mathrm{final}}$
and obtain the first-token distribution $p_1$.
Initialize the output sequence $Y$ and binding-score history
$\mathcal{H}$ as empty.\par

\medskip
\textbf{Stage 2: Decoding with BGR and PMR}\par
For each generation step $t=1,\ldots,T$, perform Steps 4--7.\par

\textbf{4. Check the greedy token's binding score.}
Set $g_t\gets\arg\max_v p_t(v)$ and compute its IVBS $b_t(g_t)$
using Algorithm~\ref{alg:ivbs}. If $\mathcal{H}$ is nonempty,
let $\mu_t=\operatorname{mean}(\mathcal{H})$.
Trigger BGR only when
\[
    \mathcal{H}\neq\emptyset
    \quad\text{and}\quad
    b_t(g_t)<r_{\mathrm{trig}}\mu_t.
\]
If BGR is not triggered, set $y_t\gets g_t$ and proceed to Step 6.\par

\textbf{5. Rethink the token choice (BGR).}
Exclude $g_t$ from the top-$k$ candidates, compute IVBS for each
remaining candidate, and retain those above the rejection threshold:
\[
    \mathcal{C}_t
    = \operatorname{TopKTokens}(p_t,k)\setminus\{g_t\},
    \qquad
    \mathcal{C}_t^{+}
    = \{y\in\mathcal{C}_t:b_t(y)\ge r_{\mathrm{rej}}\mu_t\}.
\]
If $\mathcal{C}_t^{+}$ is empty, restore $\mathcal{C}_t^{+}\gets\mathcal{C}_t$.
Select the replacement by combining log probability and IVBS, with
each score standardized over $\mathcal{C}_t^{+}$:
\[
    y_t \gets
    \arg\max_{y\in\mathcal{C}_t^{+}}
    \left[z\!\left(\log p_t(y)\right)
    +w\,z\!\left(b_t(y)\right)\right].
\]

\textbf{6. Commit the selected token.}
Append $y_t$ to $Y$ and its binding score $b_t(y_t)$ to $\mathcal{H}$.
Stop decoding if $y_t=\mathrm{EOS}$ or $t=T$.\par

\textbf{7. Reinject memory after replacement (PMR).}
Process $y_t$ and obtain its last-position hidden state $\mathbf{h}_t$
after the final RMSNorm. If BGR replaced the greedy token, mix the cached
memory into $\mathbf{h}_t$ before computing the next-token distribution:
\[
    \mathbf{h}'_t =
    \begin{cases}
        \operatorname{Mix}_{\rho}(\mathbf{h}_t,\mathbf{M}),
            & y_t\neq g_t,\\
        \mathbf{h}_t, & y_t=g_t,
    \end{cases}
    \qquad
    p_{t+1} =
    \operatorname{softmax}\!\left(
        \operatorname{Head}(\mathbf{h}'_t)\right).
\]

\textbf{return} $Y$.
\end{minipage}
\end{algorithm*}

\section{Case Study}
We compare captions generated by regular decoding and ReBind for the same images, highlighting hallucinated object mentions in red. These examples illustrate how ReBind reduces object hallucinations while preserving caption coherence.

\begin{figure*}[t] 
    \centering
    \includegraphics[width=1.0\textwidth]{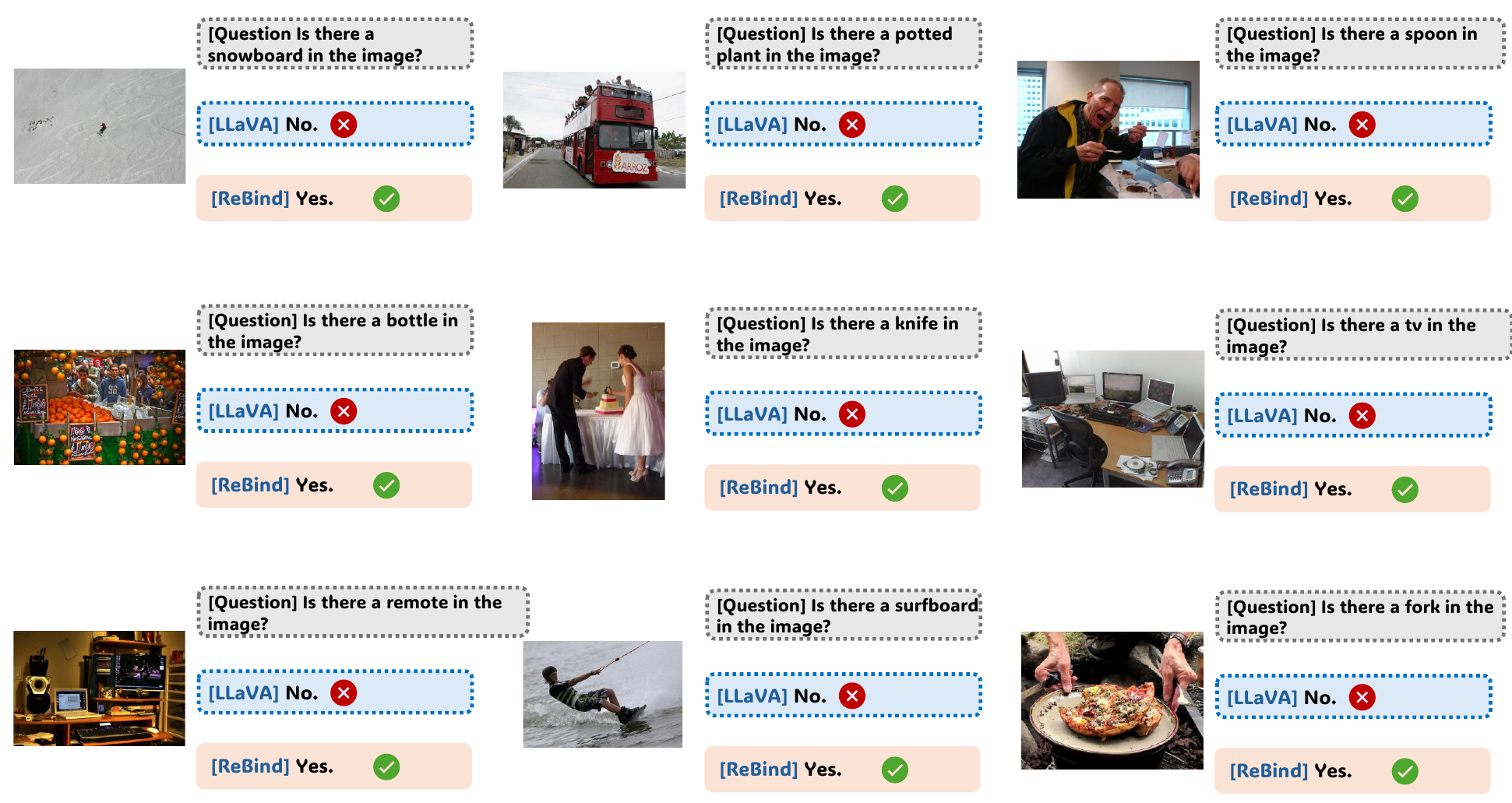}
    \caption{
Additional qualitative comparisons between vanilla LLaVA decoding and ReBind on POPE.
}
    \label{case_pope}
\end{figure*}

\begin{figure*}[t] 
    \centering
    \includegraphics[width=1.0\textwidth]{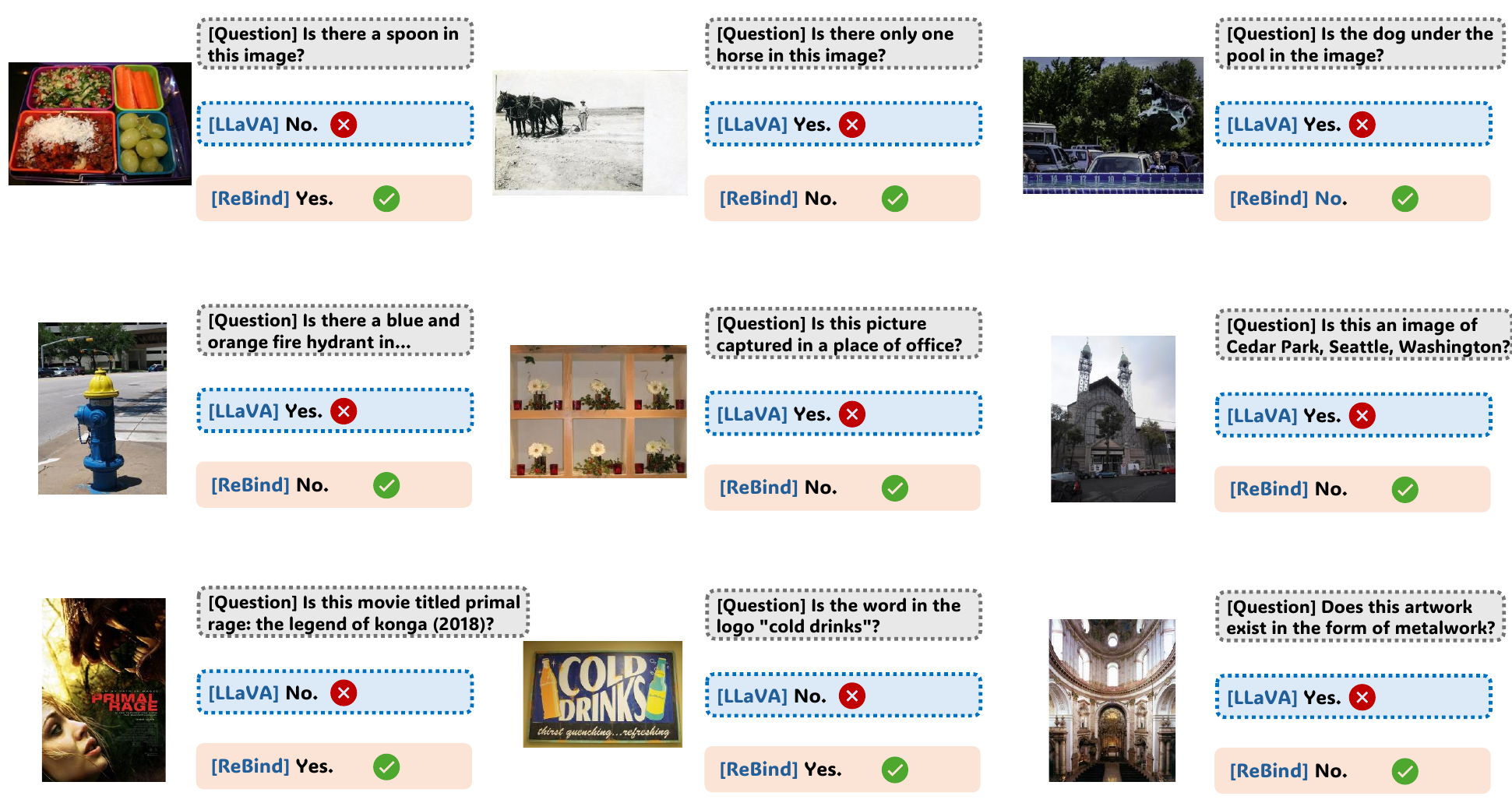}
    \caption{
Additional qualitative comparisons between vanilla LLaVA decoding and ReBind on MME.
}
    \label{case_mme}
\end{figure*}

\begin{figure*}[!t]
    \centering
    \includegraphics[width=1.0\textwidth]{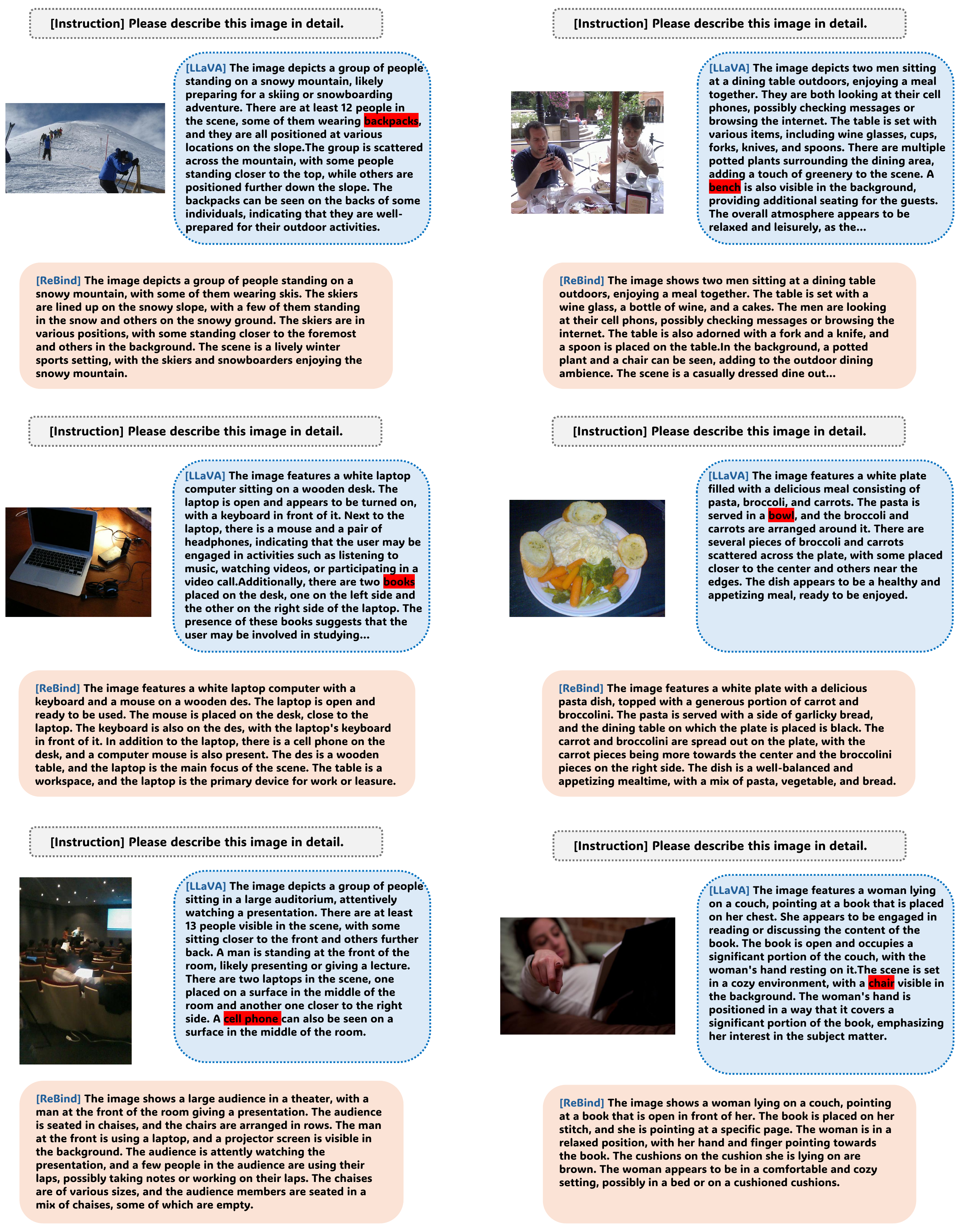}
    \caption{
    Additional qualitative comparisons between vanilla LLaVA decoding
    and ReBind on CHAIR.
    }
    \label{case_chair}
\end{figure*}

\begin{figure*}[!t]
    \centering
    \includegraphics[width=1.0\textwidth]{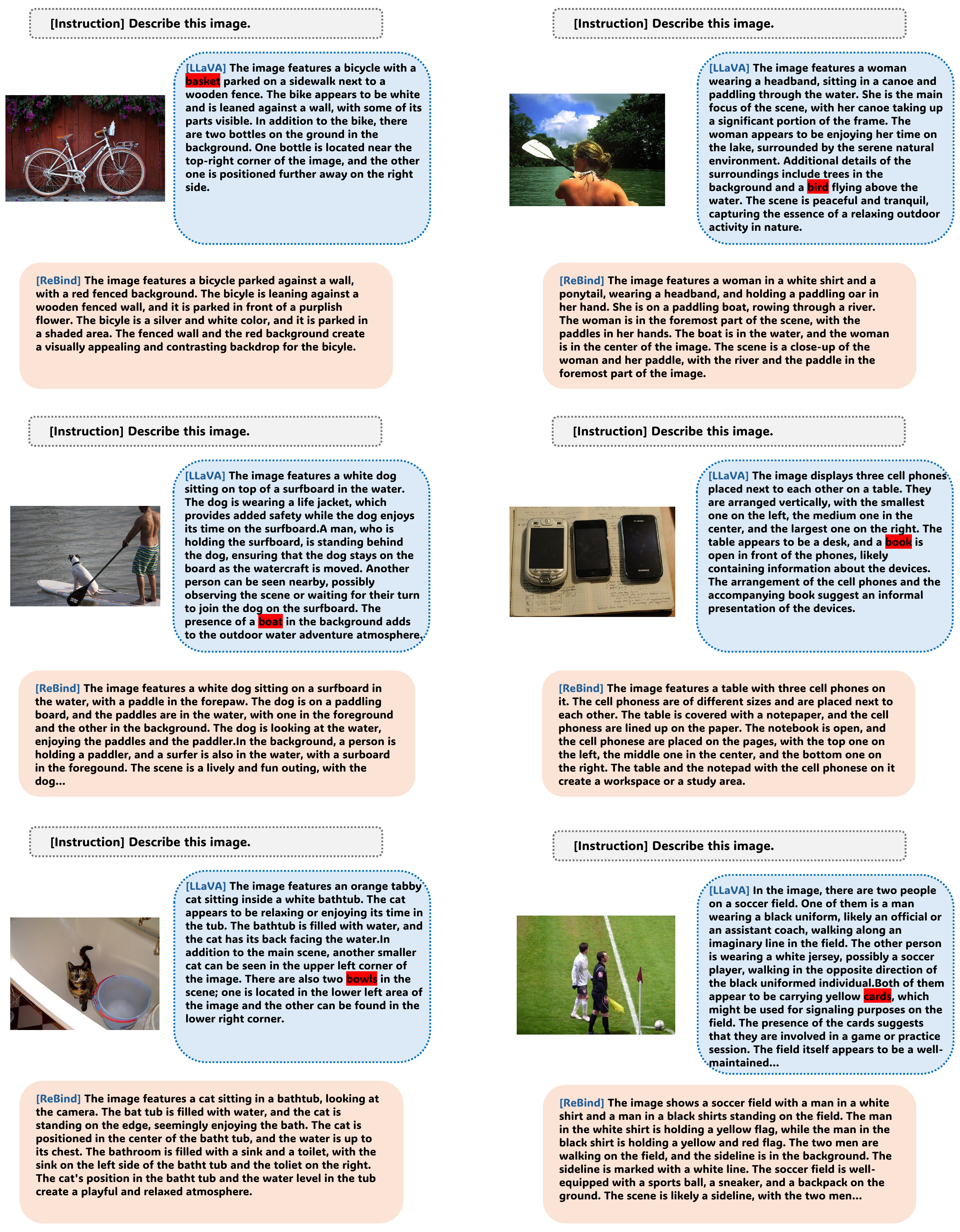}
    \caption{
    Additional qualitative comparisons between vanilla LLaVA decoding
    and ReBind on AMBER.
    }
    \label{case_amber}
\end{figure*}

\end{document}